\documentclass[10pt,twocolumn,letterpaper, table]{article}

\usepackage{cvpr}              % To produce the CAMERA-READY version
\usepackage[dvipsnames]{xcolor}

\usepackage{algorithm}
\usepackage{algpseudocode}
\usepackage{soul}
\usepackage{amsmath, amssymb, amsthm}
\usepackage{pifont}
\usepackage{multirow}
\usepackage[accsupp]{axessibility} % Improves PDF readability for those with disabilities.

\definecolor{cvprblue}{rgb}{0.21,0.49,0.74}
\usepackage[pagebackref,breaklinks,colorlinks,citecolor=cvprblue]{hyperref}

\def\paperID{3222} % *** Enter the Paper ID here
\def\confName{CVPR}
\def\confYear{2024}

\DeclareMathOperator*{\argmax}{arg\,max}
\DeclareMathOperator*{\argmin}{arg\,min}

\title{KITRO: Refining Human Mesh by 2D Clues and Kinematic-tree Rotation}

\author{Fengyuan Yang \qquad Kerui Gu \qquad Angela Yao\\
National University of Singapore\\
{\tt\small \{fyang, keruigu, ayao\}@comp.nus.edu.sg}
}

\begin{document}
\maketitle
\begin{abstract}
2D keypoints are commonly used as an additional cue to refine estimated 3D human meshes.  Current methods optimize the pose and shape parameters with a reprojection loss on the provided 2D keypoints. Such an approach, while simple and intuitive, has limited effectiveness because the optimal solution is hard to find in ambiguous parameter space and may sacrifice depth. Additionally, divergent gradients from distal joints complicate and deviate the refinement of proximal joints in the kinematic chain. To address these, we introduce \textbf{Ki}nematic-\textbf{T}ree \textbf{Ro}tation (\textbf{KITRO}), a novel mesh refinement strategy that explicitly models depth and human kinematic-tree structure. KITRO treats refinement from a bone-wise perspective.  Unlike previous methods which perform gradient-based optimizations, our method calculates bone directions in closed form.  By accounting for the 2D pose, bone length, and parent joint's depth, the calculation results in two possible directions for each child joint. We then use a decision tree to trace binary choices for all bones along the human skeleton's kinematic-tree to select the most probable hypothesis. Our experiments across various datasets and baseline models demonstrate that KITRO significantly improves 3D joint estimation accuracy and achieves an ideal 2D fit simultaneously. Our code available at: \url{https://github.com/MartaYang/KITRO}.

\end{abstract}    
\section{Introduction}
\label{sec:intro}

\begin{figure}[ht]
    % \vspace{-0.35cm}
    \centering
    \begin{subfigure}[b]{0.41\textwidth}
        \centering
        \includegraphics[width=\textwidth]{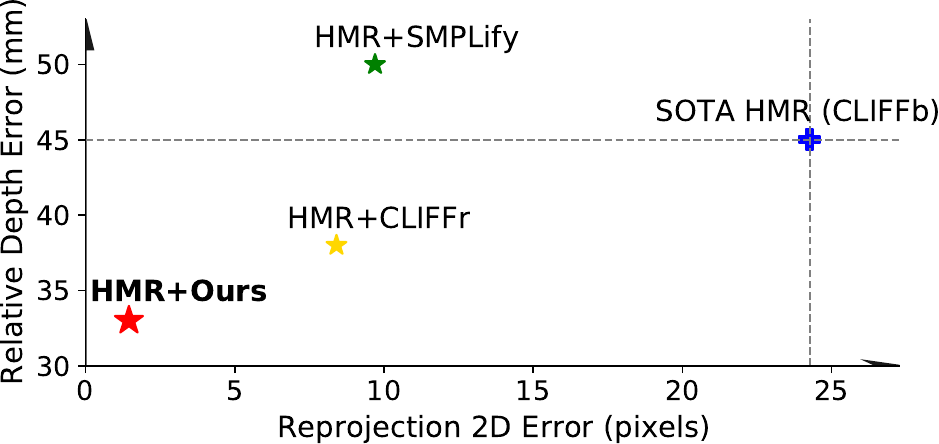}
        \vspace{-0.5cm}
        \caption{ Reprojection and depth error after 2D keypoint refinement~\footnotemark. } 
        \label{fig:teaser:UV_Z}
    \end{subfigure}
    \vspace{0.2cm} \\
    \begin{subfigure}[b]{0.41\textwidth}
        \centering
        \includegraphics[width=\textwidth]{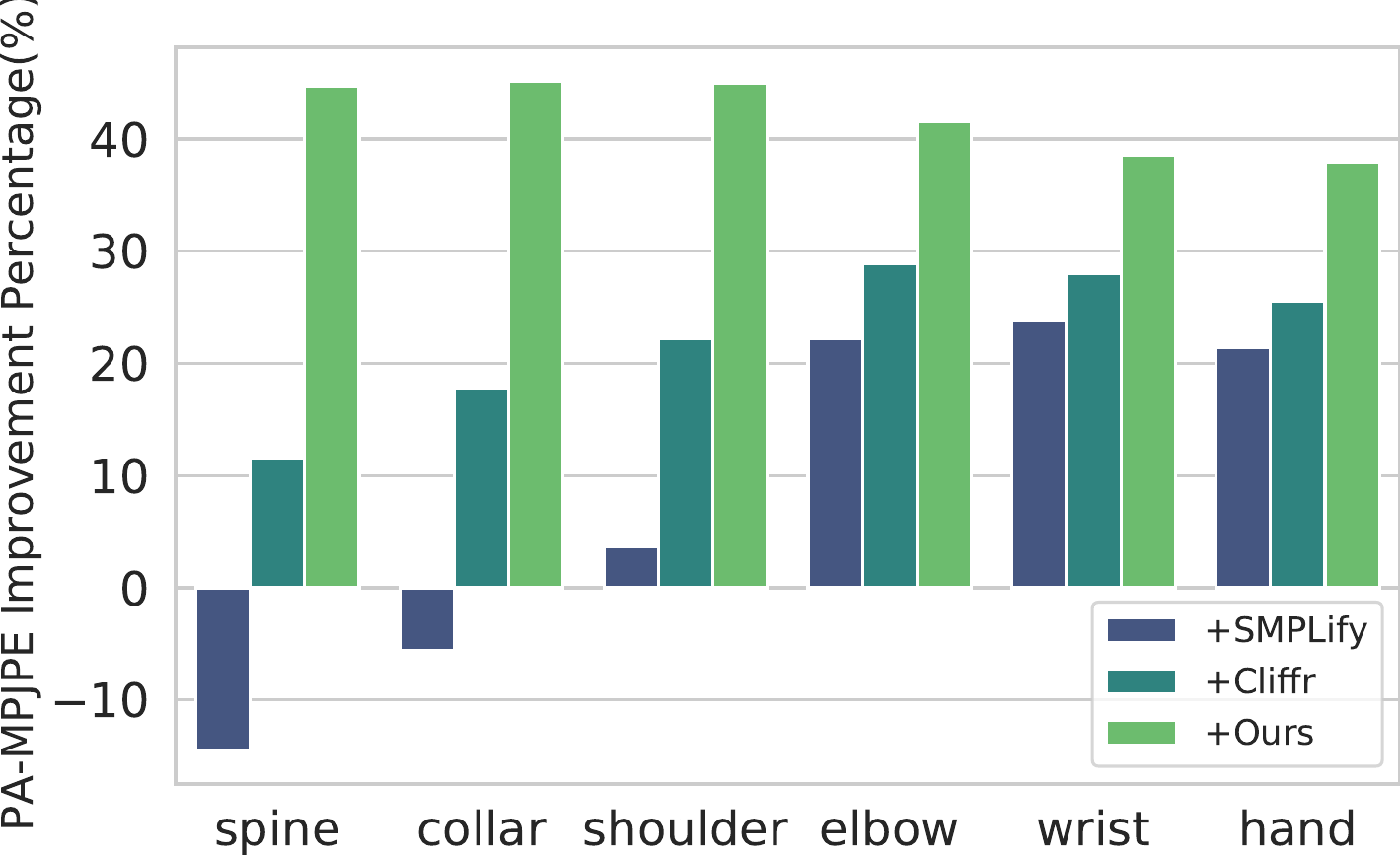}
        \vspace{-0.5cm}
        \caption{Improvement of joints along the kinematic chain of the arm.}
        \label{fig:teaser:improv_joints}
    \end{subfigure}
    \vspace{-0.25cm}
    \caption{ (a) Our method has the lowest reprojection and depth errors. 
    (b) Competing methods exhibit a decrease in improvement and even deterioration progressing along the kinematic chain from the hand to the spine; our approach exhibits the opposite but maintains a significant and competitive improvement for all joints. 
    }
    \label{fig:teaser}
    \vspace{-0.4cm}
\end{figure}

%%%%%%%%%%%%%%%%%%%%%%%%%%%%%%%%%%%%%%%%%%%%%%%%%%%%%%%%%%%%%%%%%%%%%%%%%%%%%%%%
%%%%%%%%%% Part #1: Why need 2D keypoints in human mesh recovery -- the misalignment problem
%%%%%%%%%%%%%%%%%%%%%%%%%%%%%%%%%%%%%%%%%%%%%%%%%%%%%%%%%%%%%%%%%%%%%%%%%%%%%%%%
3D human pose and shape estimation, also called Human Mesh Recovery (HMR), is relevant for augmented and virtual reality applications. Statistical human shape models like SMPL~\cite{SMPL_2015} have greatly simplified the HMR task.
However, state-of-the-art methods~\cite{EFT_2021,PARE_2021,CLIFF_2022}
 
still suffer from the ``misalignment problem"~\cite{SPIN_2019} where the predicted 3D mesh does not align well with the 2D image evidence.  The most advanced HMR model~\cite{CLIFF_2022} still has a 24 pixels reprojection error on the 3DPW dataset~\cite{3DPW_2018}, as shown in Fig.~\ref{fig:teaser:UV_Z}. On the other hand, estimating 2D human pose is more advanced~\cite{OpenPose_2017,HRNet_2020,vitpose_2022} and yields robust 2D keypoints, even under challenging scenarios such as occlusion, varying lighting, and extreme poses. 
As such, a standard strategy for improving HMR is to leverage 2D keypoints as a cue to refine the estimated 3D meshes. 
Prior methods~\cite{SMPLify_2016,EFT_2021,CLIFF_2022} optimize the pose and shape parameters with a reprojection loss on the 2D keypoints with a standard gradient descent. Such an approach, while intuitive and straightforward, is not always effective. 
\footnotetext{CLIFF~\cite{CLIFF_2022} proposes both a base model and a 2D keypoint refinement; we distinguish the two as 
`CLIFFb' and `CLIFFr'.}

%%%%%%%%%%%%%%%%%%%%%%%%%%%%%%%%%%%%%%%%%%%%%%%%%%%%%%%%%%%%%%%%%%%%%%%%%%%%%%%%
%%%%%%%%%% Part #2: The 2 problems of the previous method and our motivation.
%%%%%%%%%%%%%%%%%%%%%%%%%%%%%%%%%%%%%%%%%%%%%%%%%%%%%%%%%%%%%%%%%%%%%%%%%%%%%%%%
% ~\AY{what is the problem of the existing methods}
% ~\AY{what motivates you to develop your current methods
% previous methods solve for an estimate of the 3D pose(? theta) in an optimization approach that minimizes some objective, vs. our approach that solves for a discrete solution? 
% }

A primary reason is the inherent depth ambiguity in 2D projection, as 
multiple poses and shapes can fit the same 2D evidence. As such, optimizing solely on 2D reprojection does not account for depth ambiguity so is unlikely to find the optimal solution and may even increase the depth error (see HMR+SMPLify~\cite{SMPLify_2016} in Fig.~\ref{fig:teaser:UV_Z}). Secondly, existing methods optimize all the body joints collectively through gradient descent. Yet the gradient updates at different joints may be incongruent. Updates at the distal joints far down the kinematic chain, such as the wrist or hands, are backpropagated to the proximal joints closer to the root, like the shoulder or collar.  
Divergent gradients can complicate the update of proximal joints, limiting the refinement improvements, to the extent that it may even harm the overall accuracy (see Fig.~\ref{fig:teaser:improv_joints}).

In this work, we offer the key insight that joint depth can be solved explicitly 
in closed-form.
Provided with 2D keypoints, the depth of the parent joint, and the 3D length of the bone connecting the two joints, the problem can be narrowed down into two solutions. %  bone length, and the depth of its parent joint. 
Of these two solutions, one corresponds to the bone pointing toward the camera, and the other corresponds to the bone pointing away from the camera %The two solutions each correspond to a \hl{bone direction pointing} either towards or away from the camera shown in 
as shown in Fig.~\ref{fig:solutions}. In this way, the depth ambiguity in the solution space can be largely reduced.

Having two possible depths for every joint naturally forms a binary tree 
progressing along %the human kinematic tree 
kinematic chains in the human body (see Fig.~\ref{fig:decisiontree}). 
Any path traversing the tree is a hypothesis and we can formulate overall pose refinement as a selection problem of finding an optimal path. 
Such a strategy has a distinct advantage in that it can equally improve proximal and distal joints.

Based on these insights, we propose a novel plug-and-play %optimization-based 
human mesh refinement strategy which we call Kinematic-tree Rotation (KITRO). %~\KG{the name I feel strange}
KITRO explicitly models joint depth and solves for bone direction as a swing rotation in closed form, ensuring an excellent fit to 2D keypoints with lower depth errors. KITRO employs a decision tree to handle divergences among joints in the kinematic-tree, effectively tracing and selecting the most probable hypotheses for stable and consistent improvements across the kinematic chain. Our experimental results, on various datasets and base models, demonstrate significantly higher accuracy compared to existing methods.

%%%%%%%%%% List of contributions: 

We highlight our key contributions as follows:
\begin{itemize}
    \item KITRO's explicit depth modeling and closed-form calculation diminish ambiguities in solution space, enhancing depth accuracy and obtaining ideal 2D fit simultaneously.
    \item KITRO's decision-tree-based hypotheses tracing and selection for joint rotations encompasses the entire kinematic-tree, benefiting both proximal and distal joints.
    \item Our method's effectiveness, especially large gains in pose accuracy, is validated across datasets and base models.
\end{itemize}

\section{Related Works}
\label{sec:relatedworks}

%-------------------------------------------------------------------------
\indent \indent
\textbf{Human Mesh Recovery (HMR).}
HMR methods are either non-parametric, in that they directly estimate the 3D mesh vertices~\cite{BodyNet_2018,DenseBody_2019,I2l-meshnet_2020,pifuhd_2020,lintransformers_2021,Meshgraphormer_2021} or parametric, using statistical models~\cite{HMR_2017,SPIN_2019,rong2019delving,sengupta2020synthetic,CLIFF_2022,ImpHMR_2023}. In using a statistical model like SMPL~\cite{SMPL_2015}, the HMR task simplifies to an estimation on the model parameters.  However, estimating the parameters can be highly challenging under monocular settings despite the advancements in network architecture~\cite{GCN_hourglass_2021,HRNet_2019,transformer_2017} and learning paradigms~\cite{EvolutionaryTrain_2020,NeuralAnnot_2020,muller2021self}. 
More recently, there are also hybrid approaches combining parametric and non-parametric~\cite{Hybrik_2021}.

%-------------------------------------------------------------------------
\textbf{Human Mesh Refinement with 2D Keypoints.}
One curious observation is that estimated meshes are often poorly aligned to the 2D image evidence. As such, optimization methods have been proposed %~\cite{SMPLify_2016,SPIN_2019,EFT_2021,Hybrik_2021} 
to leverage 2D keypoints as an additional cue to refine estimated 3D human meshes.  A simple and intuitive way for refinement is simply to update the mesh model parameters, \ie the SMPL parameters with an additional 2D reprojection loss~\cite{SMPLify_2016,SPIN_2019, gu2023calibration}.
However, as 2D image evidence is an ambiguous cue for 3D estimates, directly update the SMPL parameters tends to result in unnatural poses, even if they are better aligned to the provided 2D keypoints.  As such,  
methods like EFT~\cite{EFT_2021} and CLIFF~\cite{CLIFF_2022} use the 2D reprojection loss to update the HMR estimation network weights %rather than the regression estimation network itself rather fine-tune the weights of the regression network 
instead. Such a refinement, however, is dataset-specific and thus loses generalization. Additionally, finding ways to avoid overfitting or underfitting remains a challenging limitation in these fine-tuning methods.
Our work also focuses on mesh refinement; however, instead of optimization with a 2D reprojection loss, we opt to solve for the refined solution in closed form.

\indent \indent

\section{Preliminaries}
\label{sec:Preliminaries}
% \subsection{Human Mesh Models}

\indent \indent
\textbf{SMPL.}
The SMPL model~\cite{SMPL_2015} is a commonly used 3D statistical shape model of the human body. It maps pose parameters $\boldsymbol{\theta} \in \mathbb{R}^{P \times 3}$ and shape parameters $\boldsymbol{\beta} \in \mathbb{R}^{10}$ to a 3D mesh of the body \( \mathbf{V}\!\in\!\mathbb{R}^{N \times 3} \), where \( N\!=\!6890 \) and \( P\!=\!24 \) are the number of vertices and body joints respectively. We denote the $\mathbf{J}(\boldsymbol{\beta}, \boldsymbol{\theta}) \in \mathbb{R}^{P \times 3}$ as the function that maps pose parameters $\boldsymbol{\theta}$ and shape parameters $\boldsymbol{\beta}$ to joint locations. %The mapping from $\boldsymbol{\theta}$ and $\boldsymbol{\beta}$ to 3D joints $\mathbf{J} \in \mathbb{R}^{P \times 3}$ is given by a~\AY{linear blend skinning function $W$}, \ie $\mathbf{J} = W(\boldsymbol{\theta},\boldsymbol{\beta})$.
According to human kinematic-tree shown in Fig.~\ref{fig:decisiontree}, we define the Euclidean distance between child joint $ \mathbf{J}^c(\boldsymbol{\beta}, \boldsymbol{\theta})$ and its parent joint $ \mathbf{J}^p(\boldsymbol{\beta}, \boldsymbol{\theta})$ as the bone length:
\begin{equation} \label{eq:bonelen3d}
    bl_{3D}^{(p,c)} = \left\| \mathbf{J}^{p}(\boldsymbol{\beta}, \boldsymbol{\theta}) - \mathbf{J}^{c}(\boldsymbol{\beta}, \boldsymbol{\theta}) \right\|_2.
\end{equation}

\textbf{Swing-Twist Decomposition.} The pose parameters $\boldsymbol{\theta}$ are the relative rotation of each joint, where the rotation of the pelvis~\footnote{Also known as root joint, we use them interchangeably in this paper.} joint $\boldsymbol{\theta}^0$ serves as the global rotation of the human body. %~\AY{pelvis? joint serves as the global rotation of the human body.}.  %$P$ joint rotations for $P$ joints, where 
% ~\AY{the 6D representaiton to represent the rotation of each joint, and rotates the joints along the kinematic chain from the pelvis to the hands.}
The standard SMPL model uses the axis-angle $\{\boldsymbol{\theta}^0, \cdots, \boldsymbol{\theta}^{23}\}$ to represent the rotation of each joint and rotates the joints along the kinematic chain from the pelvis to the end joints. An alternative~\cite{Hybrik_2021} proposes to represent each joint rotation in rotation matrix form $\boldsymbol{\theta}_R^i \in \mathbb{S O}(3)$ with the swing-twist decomposition: $\boldsymbol{\theta}_R^i = R_{s w} R_{t w}$, where $\boldsymbol{\theta}_R^i$ is equivalent rotation matrix form of $\boldsymbol{\theta}^i$.
Given the bone direction, the swing rotation which is 2 degrees of freedom can be derived in closed form from Rodrigues' formula (full details in Sec. \textcolor{red}{A} of the Supplementary).

\textbf{Human Mesh Recovery and Refinement.}
Monocular HMR models take an image $\mathbf{X} \in \mathbb{R}^{H \times W}$ as input and predicts SMPL parameters $({\boldsymbol{\theta}}, {\boldsymbol{\beta}})$ and the camera translation \( {\mathbf{t}} \in \mathbb{R}^{3} \) 
 %outputs the predicted SMPL parameters, as well as the 
% camera parameters translation related to human mesh \( \mathbf{t} \in \mathbb{R}^{3} \) which 
as part of the camera extrinsics to project the 3D mesh onto the image plane.
Mesh refinement methods~\cite{SMPLify_2016,EFT_2021,CLIFF_2022} uses additional 2D keypoints $\mathbf{j} \in \mathbb{R}^{P \times 2}$ to refine some estimated SMPL and camera translation parameters $(\hat{\boldsymbol{\theta}}, \hat{\boldsymbol{\beta}}, \hat{\mathbf{t}})$. The refinement process can be formalized as:
\begin{equation}\label{eq:refinedef}
\left(\hat{\boldsymbol{\theta}}',\hat{\boldsymbol{\beta}}', \hat{\mathbf{t}}' \right) = \text{Refine} (\hat{\boldsymbol{\theta}},\hat{\boldsymbol{\beta}}, \hat{\mathbf{t}}, \mathbf{j}).
\end{equation}

\noindent The refinement as given in Eq.~\ref{eq:refinedef} can also be applied iteratively such as SMPLify~\cite{SMPLify_2016}, by using the refined outputs $(\hat{\boldsymbol{\theta}}',\hat{\boldsymbol{\beta}}', \hat{\mathbf{t}}' )$ as the input estimates to continue the refinement process.
A commonly used approach for refinement is to minimize the 2D reprojection loss of the estimated 3D joints ${\mathbf{J}(\boldsymbol{\beta}, \boldsymbol{\theta})}$ with respect the provided 2D joints $\mathbf{j}$:
\begin{equation}\label{eq:reproj2Dloss}
\mathcal{L}_{j2D} =
\min_{\boldsymbol{\theta}, \boldsymbol{\beta}, \mathbf{t}} \left\| \pi\left(\mathbf{J}(\boldsymbol{\beta}, \boldsymbol{\theta}) + \mathbf{t}\right) - \mathbf{j} \right\|_2,
\end{equation}

\noindent where $\pi$ indicates the camera projection function. These parameters can be optimized jointly~\cite{EFT_2021,CLIFF_2022} or fix some parameters and optimize others~\cite{SMPLify_2016}. It is worth mentioning that when fixing $\boldsymbol{\theta}$ and $\boldsymbol{\beta}$, the optimal $\mathbf{t}$ can be solved in closed-form because it can be formulated as a least squares problem. 

\textbf{Camera Intrinsics and Extrinsics.} 
%As mentioned in \cref{sec:relatedworks}, we relax the weakly-perspective assumption by 
In our work, we relax the weakly-perspective assumption and adopt a full-perspective camera model like prior studies~\cite{FullPerspectiveSMPLify_2020,SPEC_2021,Zolly_2023}. Consistent with these works~\cite{SPEC_2021,CLIFF_2022}, we estimate the camera focal length as $f = \sqrt{H^2 + W^2}$. Therefore, the camera intrinsic can be formulated as:
\begin{equation}
K = \begin{bmatrix}
f & 0 & W/2 \\
0 & f & H/2 \\
0 & 0 & 1
\end{bmatrix}.
\end{equation}
Following the convention in HMR~\cite{HMR_2017,SPIN_2019,VIBE_2020,TCMR_2021,PARE_2021}, the camera rotation is set as $I$ and absorbed into the global orientation of the human (\ie, $\boldsymbol{\theta}^0$) predicted by the HMR model. Thus, the translation $\mathbf{t}$ is the only camera extrinsic that needs to be estimated.

\section{Method}
\label{sec:method}

\begin{figure}[tbp]
    \vspace{-0.3cm}
	\centering
	\includegraphics[width=1.\linewidth]{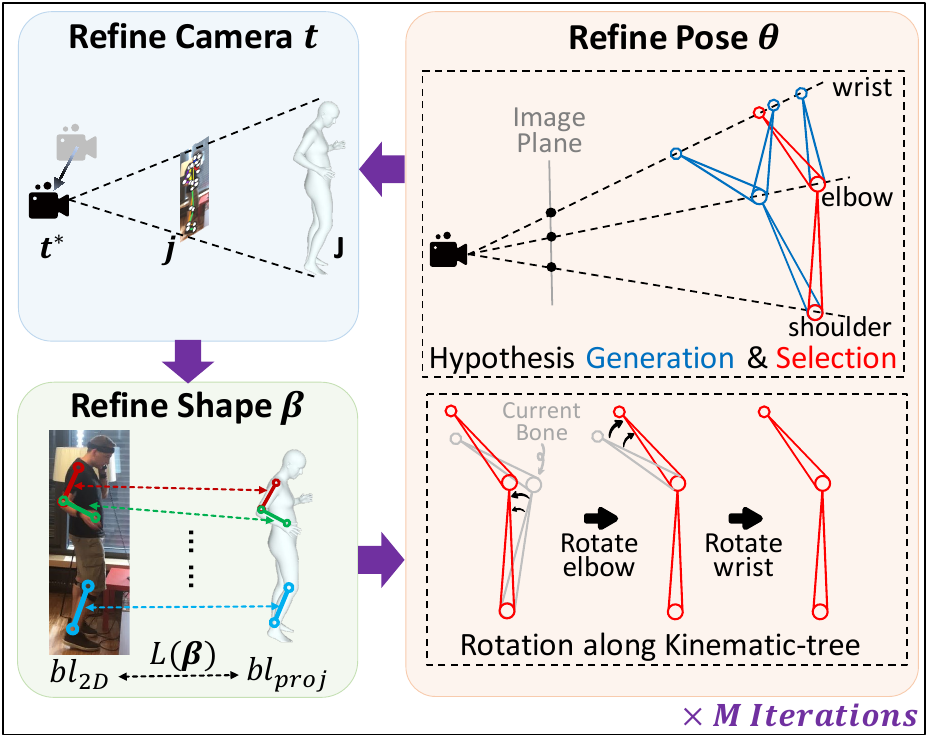}
    \vspace{-0.6cm}
	\caption{Our framework overview. Starting with an initial human mesh and 2D keypoints, our iterative refinement operates over camera, shape, and pose. The process involves localizing camera translation for 2D-3D joint alignment, optimizing shape for more aligned bone length, and generating and selecting bone direction hypotheses for rotation refinement along the kinematic-tree.
 }
	\label{fig:framework}
    \vspace{-0.3cm}
\end{figure}

As illustrated in Fig.~\ref{fig:framework}, our refinement framework iteratively updates the estimates for the camera and shape (\cref{sec:cam_shape}) as well as the pose (\cref{sec:pose}, \cref{sec:posejoint}). Unlike previous works~\cite{SMPLify_2016,EFT_2021,CLIFF_2022}, which update pose and shape parameters jointly, we perform an individual update while keeping the others fixed based on the previously refined value.  This allows us to focus on the pose refinement where we explicitly model the depth in a closed-form manner and choose the most likely hypothesis by using a decision tree along the kinematic-tree.

%-------------------------------------------------------------------------
\subsection{Camera and Shape Refinement}
\label{sec:cam_shape}
As can be seen in Fig.~\ref{fig:framework}, the given 2D keypoints together with the current human mesh can definitely provide some information and constraints for the camera translation and human shape. These constraints can help us to refine these two factors better.

%-------------------------------------------------------------------------

\textbf{Camera Translation Adjustment.}  We estimate the camera translation based on the 2D reprojection loss given by Eq.~\ref{eq:reproj2Dloss}. Minimizing the loss is equivalent to solving a least-squares optimization for the camera translation~\cite{SPIN_2019}:

\begin{equation}
\label{eq:camestimate}
    \mathbf{t}^* = \argmin_{\mathbf{t}} %\sum_{k=1}^{P} 
    \left\| \pi(\mathbf{J}(\boldsymbol{\beta},\boldsymbol{\theta}) + \mathbf{t}) - \mathbf{j} \right\|_2,
\end{equation}

\noindent where $\mathbf{t}^*$ can be found with SVD.  Note that $\mathbf{t}^*$ is the optimal solution for a given $\boldsymbol{\theta}$ and $\boldsymbol{\beta}$.

\noindent From an iterative update perspective, we find it more effective to update $\mathbf{t}$ with a moving average.  %For iteration $m$ with 
For current camera translation $\mathbf{t}$, the updated $\mathbf{t}'$ is given as:
\begin{equation}
\label{eq:cammovingavg}
    \mathbf{t}' = (\mathbf{t}^{*} + \mathbf{t}) / 2.
\end{equation}
\noindent
The reason for Eq.~\ref{eq:cammovingavg} is that $\mathbf{t}^*$ can be affected by the noise of $\boldsymbol{\theta}$ and $\boldsymbol{\beta}$. In that sense, the moving average keeps the historical information from the original HMR prediction, acting like a good regularizer.

%-------------------------------------------------------------------------
\textbf{Shape Optimization.}
% \label{sec:shape}
Previous works update $\boldsymbol{\beta}$ based on the 2D reprojection of the individual joints $\mathbf{J}(\boldsymbol{\beta},\boldsymbol{\theta})$.  We consider the refinement of $\boldsymbol{\beta}$ from a \emph{bone length} perspective. 
Consider a parent joint indexed by $p$, with estimated 3D joint $\mathbf{J}^p(\boldsymbol{\beta},\boldsymbol{\theta})$ and provided 2D joint $\mathbf{j}^p$; similarly, for a child joint indexed by $c$, consider $\mathbf{J}^c(\boldsymbol{\beta},\boldsymbol{\theta})$.
The projected 2D bone length between the $(p,c)$ joint pair is defined as the Euclidean distance between the two joints: 
\begin{equation}\label{eq:bonelengthproj}
 bl^{(p,c)}_{proj}(\boldsymbol{\beta}) = \left\| \pi(\mathbf{J}^p(\boldsymbol{\beta},\boldsymbol{\theta}) + \mathbf{t}')- \pi(\mathbf{J}^c(\boldsymbol{\beta},\boldsymbol{\theta}) + \mathbf{t}') \right\|_2.   
\end{equation}
For clarity, we specify that the projected bone length $bl^{(p,c)}_{proj}(\boldsymbol{\beta})$ depends only on $\boldsymbol{\beta}$, as we treat $\boldsymbol{\theta}$ and $\mathbf{t}'$ as fixed constants while we update the shape parameter $\boldsymbol{\beta}$.  The bone length of the provided 2D joints can be defined similarly as 
\begin{equation}\label{eq:bonelength2D}
    bl_{2D}^{(p,c)} = \left\| \mathbf{j}^{p} - \mathbf{j}^{c} \right\|_2.
\end{equation}
To estimate the shape loss, we consider an L1-norm between the projected bone length in Eq.~\ref{eq:bonelengthproj} and the given bone lenght of Eq.~\ref{eq:bonelength2D} over all the bones or ${p,c}$ combinations in the human body:
\begin{equation}\label{eq:shapeloss}
L(\boldsymbol{\beta}) = \sum_{p,c} \left| bl^{(p,c)}_{proj}(\boldsymbol{\beta}) - bl_{2D}^{(p,c)} \right|.
\end{equation}
Unlike the camera parameter $\mathbf{t}$, the optimal shape parameter $\boldsymbol{\beta}$ cannot be solved for in closed form. As such, we estimate the updated shape parameter $\boldsymbol{\beta}'$ with gradient-descent based optimization of the loss in Eq.~\ref{eq:shapeloss}:
\begin{equation}\label{eq:shapeupdate}
\boldsymbol{\beta}' = \boldsymbol{\beta}  - \eta \nabla_{\boldsymbol{\beta}} L(\boldsymbol{\beta}) \quad \text{ for } T \text{ steps},
\end{equation}
where $\eta$ is the learning rate and optimization takes T steps.

%-------------------------------------------------------------------------

\subsection{Pose Hypothesis Generation}\label{sec:pose}

% ~\AY{intro on how the body poses are solved in closed form, form a solution set; next section, outline how we select the "optimal" solution.}

% \todo{kinematic tree}
\indent \indent
\textbf{3D Bone Direction Calculation.}
\begin{figure}[!tbp]
    \vspace{-0.3cm}
	\centering
	\includegraphics[width=0.8\linewidth]{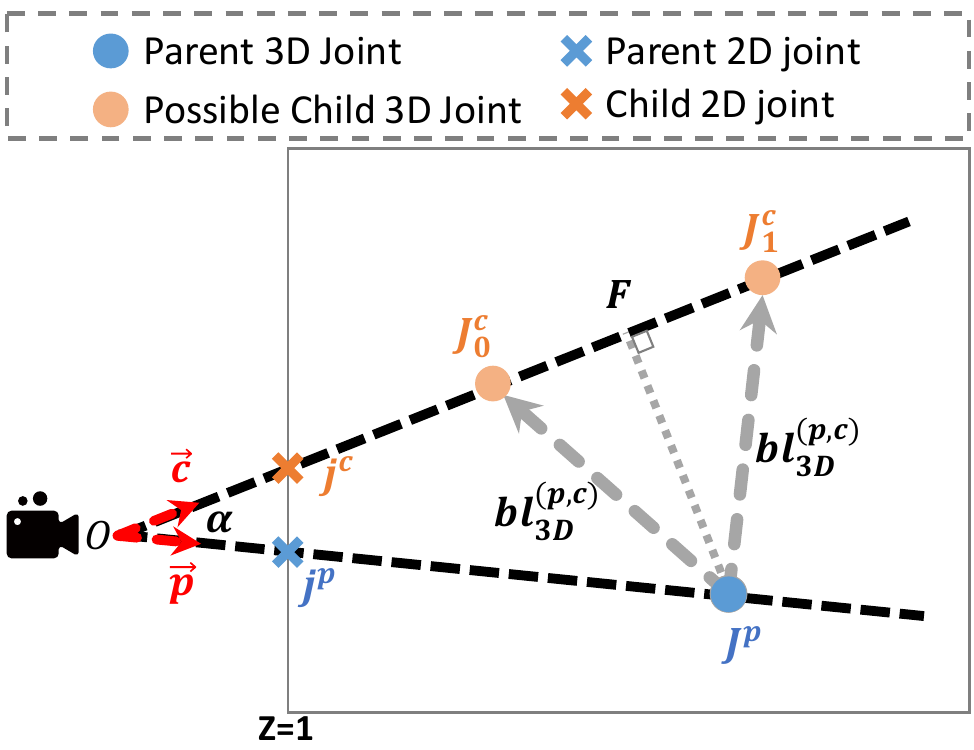}
    \vspace{-0.3cm}
	\caption{Calculation of two bone direction solutions of the (p, c) joint pair, based on 2D keypoints, bone length, and the depth of the parent joint. One points towards the camera, and the other away.}
	\label{fig:solutions}
    \vspace{-0.3cm}
\end{figure}
To refine the pose parameters $\boldsymbol{\theta}$, we opt to work with the swing-twist decomposed form of the joint rotation, rather than in the standard axis-angle representations of SMPL.  This is key in our refinement approach because it allows us to estimate the swing angle in closed form directly from the provided 2D keypoints.  Consider the 2D keypoints \( \mathbf{j}^p \) and \( \mathbf{j}^{c} \) of a bone \( (p, c) \) as defined in the SMPL kinematic tree. Their corresponding 3D joints $\mathbf{J}^p(\boldsymbol{\beta}',\boldsymbol{\theta})$ and $\mathbf{J}^c(\boldsymbol{\beta}',\boldsymbol{\theta})$ are located along the rays $\vec{p}$ and $\vec{c}$ from the camera to the 2D keypoints on the image plane. And $O$ denotes the camera location which is the negative of camera translation $O = -\mathbf{t}'$ since camera rotation is identity.
As shown in Fig.~\ref{fig:solutions}, directions of these two rays are:
\begin{equation}
    \vec{p} = K^{-1} \times \mathbf{j}^p_{h}, \quad
    \vec{c} = K^{-1} \times \mathbf{j}^c_{h},
\end{equation}
where subscript $h$ denotes homogenized coordinates and $K$ is the camera intrinsic matrix defined in \cref{sec:Preliminaries}.
Based on the direction of the rays $\vec{p}$ and $\vec{c}$, their intersection angle $\alpha$ satisfies the following:
\begin{equation}
    \cos \alpha = \frac{\vec{p} \cdot \vec{c}}{\lVert \vec{p} \rVert \lVert \vec{c} \rVert}, \quad
    \sin \alpha = \frac{\lVert \vec{p} \times \vec{c} \rVert}{\lVert \vec{p} \rVert \lVert \vec{c} \rVert}.
\end{equation}
Now, as we know the 3D bone length $bl_{3D}^{(p,c)}$ defined in Eq.~\ref{eq:bonelen3d} % $d_{3D}^{p,c} = \left\| \mathbf{J}_{i'}^p - \mathbf{J}_{i'}^c \right\|^2_2$ 
and the depth of the parent joint from the camera % \(|O {J}^p| = \left\| \mathbf{J}_{i'}^p - \mathbf{t}_{i+1} \right\|_2\), where \( \mathbf{J}_{i'} = W(\boldsymbol{\theta}_i,\boldsymbol{\beta}_{i+1}) \) 
\(|O {J}^p| = \left\| \mathbf{J}^p(\boldsymbol{\beta}', \boldsymbol{\theta}) + \mathbf{t}' \right\|_2\), we can directly solve for the child joint in closed form.  Specifically, $|F J^c_0| = |F J^c_1| = \sqrt{\left(bl_{3D}^{(p,c)}\right)^2 - \left(\left|OJ^p\right| \sin \alpha\right)^2}$ based on Fig.~\ref{fig:solutions}. Therefore, we can calculate the child joint $J^c$ specified by the vector $\overrightarrow{J^p J^c}$ from the parent joint $J^p$, and the depth $|O J^p|$ from the camera $O$:
\begin{equation}\label{eq:solutions}
\begin{aligned}
    &\overrightarrow{J^p J^c}=\underbrace{|O J^p| \cdot(\cos \alpha \cdot \vec{c}-\vec{p})}_{\overrightarrow{J^p F}} \pm \underbrace{|F J^c_0| \cdot \vec{c}}_{\overrightarrow{F J^c_0}}, \\
    &|O J^c|={|O J^p| \cdot \cos \alpha} \pm {|F J^c_0|}.
\end{aligned}
\end{equation}
There are two possibilities for the child joint, as indicated by the $\pm$ sign in the solutions in Eq.~\ref{eq:solutions}. One solution points towards the camera ($\overrightarrow{J^p J^c_0}$ in ~\cref{fig:solutions}) while the other points away ($\overrightarrow{J^p J^c_1}$).  These two solutions directly illustrate the inherent depth ambiguity. Depending on the accuracy of the estimated terms, the square root term may become negative (might arise during decision tree calculations in the next step); for numerical stability, we rectify it to 0.

\textbf{From Bone Direction to Full Body Pose Hypothesis.}
Note that the solution for the bone direction in Eq.~\ref{eq:solutions} depends on the depth of the parent joint \(|O J^p|\).  As the parent joint's depth has two possible solutions as well, and the depth dependency propagates through the kinematic tree, the hypotheses for possible poses naturally form binary trees. Fig.~\ref{fig:decisiontree} shows an example of the left leg, beginning at the pelvis root node, with child nodes for the hip, knee, ankle, and toe. For the entire body pose, there are 5 trees, representing the arms, legs, and torso. A full-body pose hypothesis is then represented by a path through each of the trees. It's worth mentioning that while we could select solutions for each joint greedily, our approach leverages a decision tree naturally formed along the kinematic chain since it enables more accurate selections by considering all potential outcomes.

\subsection{Pose Hypothesis Selection}\label{sec:posejoint}
\begin{figure}[!tbp]
    \vspace{-0.3cm}
	\centering
	\includegraphics[width=1.0\linewidth]{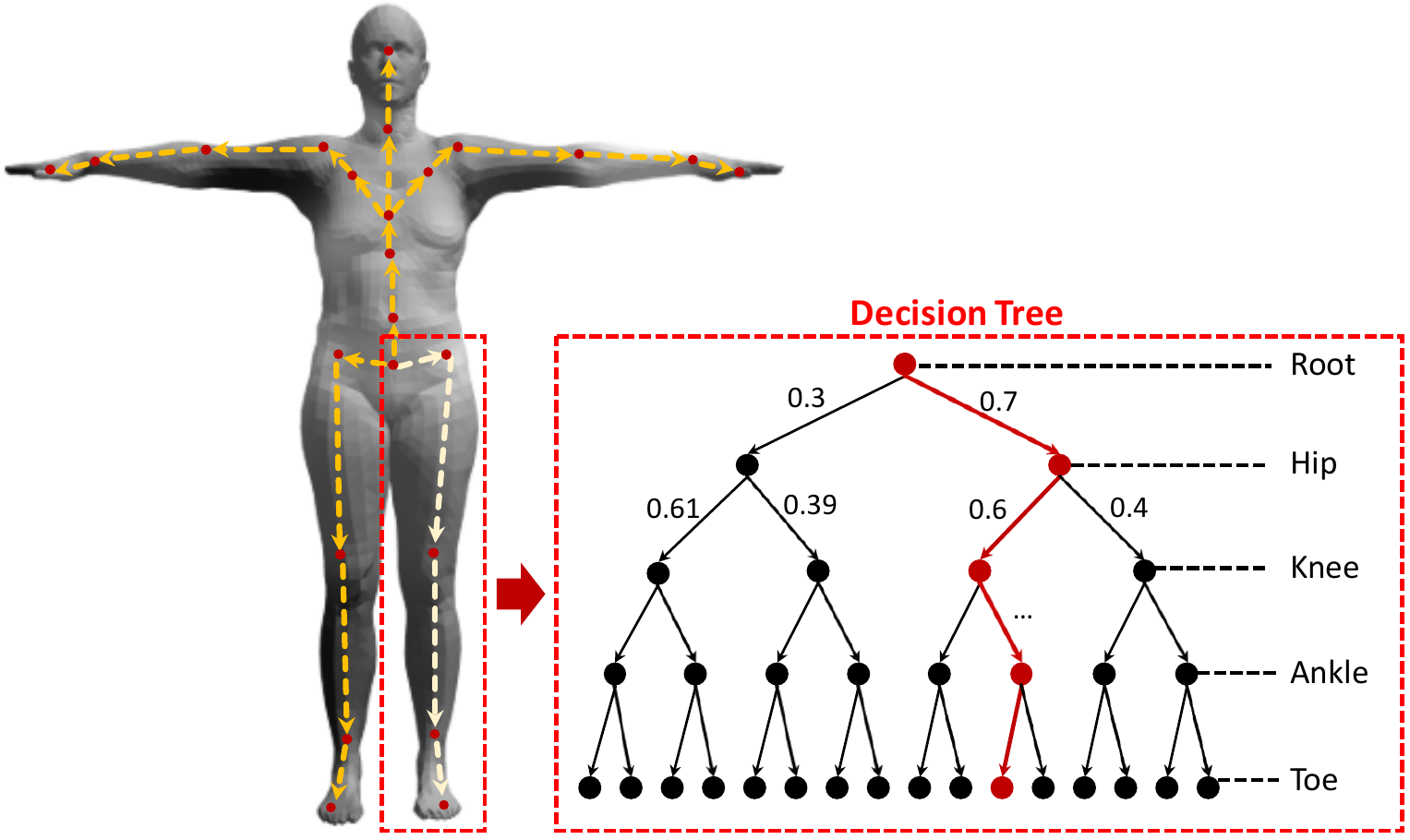}
    \vspace{-0.55cm}
	\caption{Human kinematic-tree in SMPL~\cite{SMPL_2015} and ours decision tree example tracing all hypotheses for leg joints. In our method, local solution certainties are used as edge weights, with the largest global path product representing the most probable pose.}
	\label{fig:decisiontree}
    \vspace{-0.4cm}
\end{figure}

\indent \indent
\textbf{Decision Tree Formulation.}
From the binary trees, we wish to select an optimal path.  Yet without additional information or cues, it is challenging to know how optimality should be measured.  As such, we rely only on the initial HMR estimate, and 
compute the cosine similarity between the relative bone rotations for the two closed-form solutions $R_{rel}(\overrightarrow{J^p J^c_k}| \phi(p))$ for $k=\{0, 1\}$ and the relative bone rotation predicted by the original HMR model $\boldsymbol{\theta}_\text{HMR}^p$:
% $\text{cos\_sim}(R_{rel}(\overrightarrow{J^p J^c_k}, \phi(p)), \boldsymbol{\theta}_{H M R}^c)$,
\begin{equation}
    \text{cos\_sim}(\overrightarrow{J^p J^c_k}|\phi(p)) = \frac{cos \left< R_{rel}(\overrightarrow{J^p J^c_k}| \phi(p)), \boldsymbol{\theta}_\text{HMR}^p \right> + 1}{2}
\end{equation}
where \( \phi(p) = (\phi_0, \cdots, \phi_i, \cdots, \phi_p) \) represents the path from the root to the current parent joint \(p\). Each \(\phi_i \in \{0, 1\}\) denotes the two possible solutions for each bone with $\mathbf{J}_i$ as the child joint along the chain, except \(\phi_0\) since the root joint is fixed after camera refinement for each iteration. 
We directly apply the Softmax of these two cosine similarities as the weights for edges in the decision tree as shown in Fig.~\ref{fig:decisiontree}:
\begin{equation}\label{eq:solutionselection}
    w(e^{(p,c)}_k|\phi(p)) = \frac{\exp(\text{cos\_sim}(\overrightarrow{J^p J^c_k}|\phi(p)))}{ \sum_{k'=0}^1 
    \exp(\text{cos\_sim}(\overrightarrow{J^p J^c_{k'}}|\phi(p)))}.
\end{equation}
Intuitively, Eq.~\ref{eq:solutionselection} measures the relative consistency between calculated and original prediction. It relies on the assumption that %The insight behind this practice is that: although the 
the original HMR prediction, although not precise at exact bone direction, % even if it is not precisely accurately, % is not precise on the fine-grained bone direction prediction, it is at least almost correct on the coarse-grained selection of whether 
is sufficiently accurate at indicating whether the bone points towards or away from the camera.  %the bone is pointing forward or away from the camera, which can serve as a strong prior. 
% Proof-of-concept experiments that verified this insight can be found in the supplementary materials. 
We verify this assumption empirically in Sec. \textcolor{red}{B} of the Supplementary.  

\noindent
The computational complexity for calculating the solutions in the binary trees depends only on the depth, as nodes within any given depth of the trees can be computed in parallel. 
By constructing the binary trees and estimating the weights, we obtain a %can have a more 
global view on the pose feasibility.  We choose as the optimal path the one with the highest node products. The final pose is defined by the optimal selection path:
\begin{equation}\label{eq:optimalsolution}
    \phi^* = \argmax_{\phi}{\prod_{(p,c)} w(e^{(p,c)}_{\phi_c}|\phi(p))},
\end{equation}
where $\phi = \{\phi_0, \cdots, \phi_{23}\}$ is a pose hypothesis for all 23 bones of a human body.

\textbf{Pose Parameter Update.}
We perform an update on current \(\boldsymbol{\theta}\) according to the selection chain \( \phi^* \) in Eq.~\ref{eq:optimalsolution}.  
% as the final step. To ensure that the refinement step isn't too drastic, we also adopt a re-weighting strategy for \(\boldsymbol{\theta}_i\) updating. 
The update is soft, weighted by the edge weight $w(e^{(p,c)}_{\phi^*_c}|\phi^*(p))$ calculated in Eq.~\ref{eq:solutionselection} which acts like a certainty term for choosing $\phi^*_c$ instead of the other (\ie, $1-\phi^*_c$):
\begin{equation}
    \lambda^{(p,c)} = w(e^{(p,c)}_{\phi^*_c}|\phi^*(p)).
\end{equation}
The final bone direction $\vec{\mathbf{b}}_{new}^{(p,c)}$ is updated as the weighted sum of the selected solution $\overrightarrow{{J^p J^c_{\phi_c^*}}}$ and current bone direction $\vec{\mathbf{b}}^{(p,c)} = \mathbf{J}^c(\boldsymbol{\beta}', \boldsymbol{\theta}) - \mathbf{J}^p(\boldsymbol{\beta}', \boldsymbol{\theta})$:
\begin{equation}
\label{eq:posereweight}
    \vec{\mathbf{b}}_{new}^{(p,c)} = \lambda^{(p,c)} \cdot \overrightarrow{{J^p J^c_{\phi_c^*}}} + (1-\lambda^{(p,c)}) \cdot \vec{\mathbf{b}}^{(p,c)}.
\end{equation}
Finally, the pose parameter \(\boldsymbol{\theta}^p\) of the parent joint \(p\) can be updated based on the % There are two scenarios:
swing rotation $R_{sw}^{(p,c)}$ which rotates \( \vec{\mathbf{b}}^{(p,c)} \) to \( \vec{\mathbf{b}}_{new}^{(p,c)} \) simply by the Rodrigues' rotation formula.
In the special case of the root joint and the third spine joint (\ie, `Spine3')%~\AY{is there a name in the code?}
, where there are three children denoted as $c_0,c_1,c_2$, we compute a rotation matrix \(R_{sw}^{(p,c)}\) that optimally rotates the vectors \(\{\vec{\mathbf{b}}^{(p,c_0)}, \vec{\mathbf{b}}^{(p,c_1)}, \vec{\mathbf{b}}^{(p,c_2)}\}\) to best align with \(\{\vec{\mathbf{b}}_{new}^{(p,c_0)}, \vec{\mathbf{b}}_{new}^{(p,c_1)}, \vec{\mathbf{b}}_{new}^{(p,c_2)}\}\) by SVD~\cite{Hybrik_2021}.
Finally, by applying above rotation matrix \(R_{sw}^{(p,c)}\) to update parent joint $p$'s rotation $\boldsymbol{\theta}_R^p$, we obtain the refined pose parameter:
\begin{equation} \label{eq:thetaupdating}
    {\boldsymbol{\theta}_R^p}'=(\prod_{i \in KC(\Tilde{p})} \boldsymbol{\theta}_R^i)^T \cdot R_{sw}^{(p,c)} \cdot \prod_{i \in KC(p)} \boldsymbol{\theta}_R^i,
\end{equation}
where \(\Tilde{p}\) is the parent joint of \(p\), and $KC(p)$ is the kinematic chain from root to joint $p$. Proof of correctness is given in Sec. \textcolor{red}{C} of the Supplementary. Updating all joints by Eq.~\ref{eq:thetaupdating} along kinematic-tree gets the refined $\boldsymbol{\theta}'$.

The updates in Eq.~\ref{eq:cammovingavg}, Eq.~\ref{eq:shapeupdate}, and Eq.~\ref{eq:thetaupdating} specify one refinement iteration for $\mathbf{t}$, $\boldsymbol{\beta}$ and $\boldsymbol{\theta}$. We can continue to iterate by using the refined parameters as initial estimates for a total of \(M\) iterations.

%-------------------------------------------------------------------------
\subsection{Implementation Details}
Here we elaborate on more implementation details.
For the shape refinement, we employ Adam optimizer to optimize Eq.~\ref{eq:shapeloss} for $T=10$ steps with a learning rate $\eta=0.1$ in each iteration. And the whole iteration number is $M=10$. Ours overall pseudo-code is shown in \cref{alg}.

\begin{algorithm}[!t]
\caption{Ours Human Mesh Refinement}
\begin{algorithmic}[1]
\Require Initial pose $\boldsymbol{\theta}_0$, shape $\boldsymbol{\beta}_0$, camera translation $\mathbf{t}_0$, 2D keypoints $\mathbf{j}$, iterations $M$, kinematic-tree $KT$.
\Ensure Refined $\boldsymbol{\theta}_M$, $\boldsymbol{\beta}_M$, $\mathbf{t}_M$. %Total loop number is $M$.

\For{$m = 0 \to M-1$}
\State $\mathbf{t}^* \leftarrow $ Best $\mathbf{t}$ aligned $\mathbf{j}$ and $\mathbf{J}(\boldsymbol{\beta}_m,\boldsymbol{\theta}_m)$ \Comment{Eq.~\ref{eq:camestimate}}
\State $\mathbf{t}_{m+1} \leftarrow (\mathbf{t}^{*} + \mathbf{t}_m) / 2$   \Comment{Eq.~\ref{eq:cammovingavg}}
\State $\boldsymbol{\beta}_{m+1} \leftarrow$ Adam optimize $\boldsymbol{\beta}_{m}$ by $L(\boldsymbol{\beta})$ \Comment{Eq.~\ref{eq:shapeloss}}
\State $DecisionTree \leftarrow$ Binary solutions along $KT$ %\Comment{Eq.\ref{eq:solutions}}
\State $\phi^* \leftarrow$ Optimal path in $DecisionTree$ \Comment{Eq.~\ref{eq:optimalsolution}}
\State $\boldsymbol{\theta}_{m+1} \leftarrow$ Update bone rotation base on $\phi^*$   \Comment{Eq.~\ref{eq:thetaupdating}}
\EndFor
\end{algorithmic}
\label{alg}
\end{algorithm}

 \section{Experiments}
\label{sec:Experiments}

%-------------------------------------------------------------------------
\subsection{Dataset and Metrics}

\indent \indent
\textbf{3DPW} dataset serves as a rigorous outdoor benchmark tailored for 3D pose and shape estimation. We follow~\cite{CLIFF_2022,EFT_2021,DynaBOA_2022} and use the ground truth 2D keypoint annotations from 3DPW as refinement inputs. %are utilized as the 2D indicators.
\textbf{Human3.6M} %stands as 
is an indoor 3D human mesh dataset. Consistent with preceding works~\cite{HMR_2017,SPIN_2019,pose2mesh_2020}, we use subjects S9 and S11 for testing. Similarly, we use the ground truth 2D keypoints for refinement inputs.  %annotations from this dataset are adopted as the 2D indicators.

We use three standard 3D pose and shape metrics: (1) \textbf{MPJPE} measures the average Euclidean distance between the ground truth and the predicted joint positions, only considering alignment at the pelvis, (2) \textbf{PA-MPJPE} which is the MPJPE error after further aligning the predicted pose to the ground truth with Procrustes aligned and (3) \textbf{PVE} measuring the average Euclidean distance between the predicted and the ground truth mesh vertices after pelvis alignment.  % vertex positions of the mesh. 

%-------------------------------------------------------------------------
\subsection{Ablation Study}

We adopted CLIFFb~\cite{CLIFF_2022} as the baseline HMR model. The ablation study upon our full model for three refinement factors, camera, shape, and pose, are shown in Tab.~\ref{tab:abaltion}.

\begin{table}[t!]
    \vspace{-0.3cm}
    \centering
    \caption{Ablation study for camera, shape, and pose refinement on 3DPW across three segments. The bottom segment presents our full model's results. `DT' stands for Decision Tree.}
    \vspace{-0.25cm}
    \resizebox{1.0\linewidth}{!}{
    \begin{tabular}{lccc}
        \hline
        Method & PA-MPJPE $\downarrow$ & MPJPE $\downarrow$ & PVE $\downarrow$ \\
        \hline
        %KITRO w/o CE 
        fixed $\mathbf{t}$ as $\mathbf{t}_0$ & 28.53 & 46.68 & 57.76 \\
        hard-update $\mathbf{t}$ (Eq.~\ref{eq:camestimate}) & 27.99 & 50.21 & 61.11 \\
        Large focal length & 32.72 & 52.46 & 63.61 \\
        \hline
        fixed $\boldsymbol{\beta}$ as $\boldsymbol{\beta}_0$ & 35.00 & 69.02 & 84.25 \\
        $\boldsymbol{\beta}$ from $\mathcal{L}_{j2D}$ of Eq.~\ref{eq:reproj2Dloss} & 32.60 & 52.60 & 64.45 \\
        \hline
        fixed $\boldsymbol{\theta}$ as $\boldsymbol{\theta}_0$ & 44.57 & 80.03 & 95.71  \\
        greedy + hard-update & 34.32 & 54.91 & 67.05 \\
        greedy + soft-update & 29.24 & 45.94 & 56.57 \\
        DT + hard-update  & 32.18 & 52.15 & 63.82   \\
        \hline
        \rowcolor{lightgray!30}
        KITRO (\textbf{ours}) & \textbf{27.67} & \textbf{43.53} & \textbf{53.44} \\
        \hline
    \end{tabular}}
    \label{tab:abaltion}
\end{table}

\textbf{Camera Refinement.} The first segment in Tab.~\ref{tab:abaltion} shows the impact when we do not refine the camera estimate and fix it to the HMR estimate (first row) and do not use the moving average and directly replace the camera estimate according to the optimal translation of Eq.~\ref{eq:camestimate} without Eq.~\ref{eq:cammovingavg} (second row). Performance drop shows the effectiveness of the camera translation refinement and the moving average as a good regularizer. 
The third row, using a fixed focal length of 5000, shows an obvious performance drop, indicating the limitations of a weak perspective camera model.

\textbf{Shape Refinement.} The second segment in Tab.~\ref{tab:abaltion} shows the impact when the shape parameter $\boldsymbol{\beta}$ is not updated (first row in this segment) vs. optimizing $\boldsymbol{\beta}$ according to the standard 2D projection loss of the keypoints (second row) and our proposed 
loss on the 2D bone length projection (as per Eq.\ref{eq:shapeloss}). % indeed is more suitable than the traditional reprojection loss in our methodology.
This comparison reveals that our loss is more effective than $\mathcal{L}_{j2D}$ in our method. The ablation for refinement step $T$ and learning rate $\eta$ are given in Sec. \textcolor{red}{D} of the Supplementary, they are all not highly sensitive factors.

\begin{table*}[t!]
    \vspace{-0.3cm}
    \centering
    \caption{SOTA comparison on 3DPW and Human3.6M. Baseline HMR model for human mesh initialization in the first segment, refinement methods in the second segment including our results at the bottom. $^*$ indicates our reproduced results. $^\dagger$ marks a different 3DPW evaluation protocol using extra gender information in data preprocessing. All our reproductions ensure the same protocol and fair comparison.}
    \vspace{-0.3cm}
    \resizebox{0.95\linewidth}{!}{
    \begin{tabular}{lcccccc}
    % \begin{tabular}{llllllll}
        \hline 
        & \multicolumn{3}{c}{3DPW} & & \multicolumn{2}{c}{Human3.6M} \\
        \cline { 2 - 4 } \cline { 6 - 7 }
        Method & PA-MPJPE $\downarrow$ & MPJPE $\downarrow$ & PVE $\downarrow$ & & PA-MPJPE $\downarrow$ & MPJPE $\downarrow$ \\%& PVE $\downarrow$ \\
        \hline 
        CLIFFb$^\dagger$~\cite{CLIFF_2022} & 43.0 & 69.0 & 81.2     & & - & - \\%& - \\
        CLIFFb$^*$ & 43.76 & 73.67 & 91.58     & & 36.16 & 55.18 \\%& 74.46 \\
        \hline 
        DynaBOA~\cite{DynaBOA_2022} & 40.4 & 65.5 & 82.0    & & - & - \\%& -\\
        Pose2Mesh~\cite{pose2mesh_2020} & 34.6 & 65.1 & -     & & 35.3 & 51.1 \\%& -\\
        CLIFFb + CLIFFr$^\dagger$~\cite{CLIFF_2022}  & 32.8 & 52.8 & 61.5      & & - & - \\%& -\\
        CLIFFb + SMPLify$^*$ & 36.11 & 66.67 & 79.91      & & 28.07 & 45.19 \\%& 63.08\\
        CLIFFb + CLIFFr$^*$ & 32.04 & 55.83 & 71.95      & & 25.88 & 42.79 \\%& 60.60\\
        % CLIFFb + CLIFFr (3DPW)$^*$ & 32.04 & 55.83 & 71.95      & & 36.16 & 55.18 & 74.46\\
        % CLIFFb + CLIFFr (Human3.6M)$^*$ & 46.17 & 77.85 & 94.71      & & 25.88 & 42.79 & 60.60\\
        % \hline
        \rowcolor{lightgray!30}
        CLIFFb + KITRO \textbf{(ours)}  & \textbf{27.67 \textcolor{blue}{(4.3$\downarrow$)}} & \textbf{43.53 \textcolor{blue}{(12.3$\downarrow$)}} & \textbf{53.44 \textcolor{blue}{(18.5$\downarrow$)}}      & & \textbf{21.04 \textcolor{blue}{(4.8$\downarrow$)}} & \textbf{34.50 \textcolor{blue}{(8.3$\downarrow$)}} \\%& \textbf{42.88 \textcolor{blue}{(17.7$\downarrow$)}}\\
        \hline
    \end{tabular}}
    \label{tab:SOTA}
    \vspace{-0.5cm}
\end{table*}

\textbf{Pose Refinement.} The third segment in Tab.~\ref{tab:abaltion} shows the impact when the pose refinement is removed (first row in this segment) as well as alternative designs on the decision tree (vs greedy selection, second and third row) and the weighted update of Eq.~\ref{eq:posereweight} (vs hard-update, second and fourth row). The decreased performance in these designs shows the global perspective offered by our decision tree and the necessity of reweighting for a moderated update.

\textbf{Parameter Impact.} Across the three refined parameters, $\boldsymbol{\theta}$ has the biggest impact; removing it from the refinement leads to a $\approx$60\% increase in these three evaluation errors. In contrast, $\boldsymbol{\beta}$ has a $\approx$30\% increase, and the camera has the least impact of $<1\%$. This demonstrates the crucial of pose and shape refinement in our method, with the KITRO design for pose refinement notably enhancing performance. Camera refinement, however, exhibits less sensitivity.

\begin{figure}[!tbp]
    \vspace{-0.4cm}
	\centering
	\includegraphics[width=1.0\linewidth]{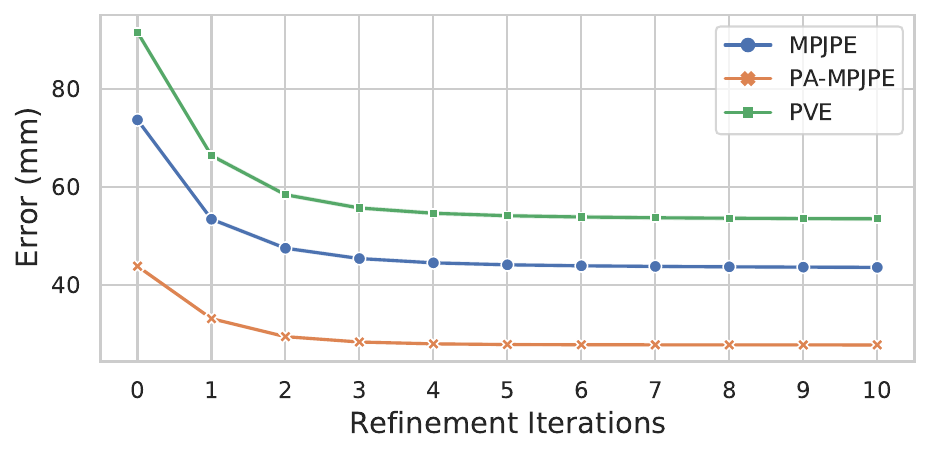}
    \vspace{-0.8cm}
	\caption{Impact of Refinement Iterations for MPJPE, PA-MPJPE, and MPVPE on 3DPW. Three line plots demonstrate a graceful decrease and quick convergence of error metrics.}
	\label{fig:n_loop}
    \vspace{-0.55cm}
\end{figure}

\begin{figure*}[!tbp]
    \vspace{-0.55cm}
	\centering
	\includegraphics[width=1.0\linewidth]{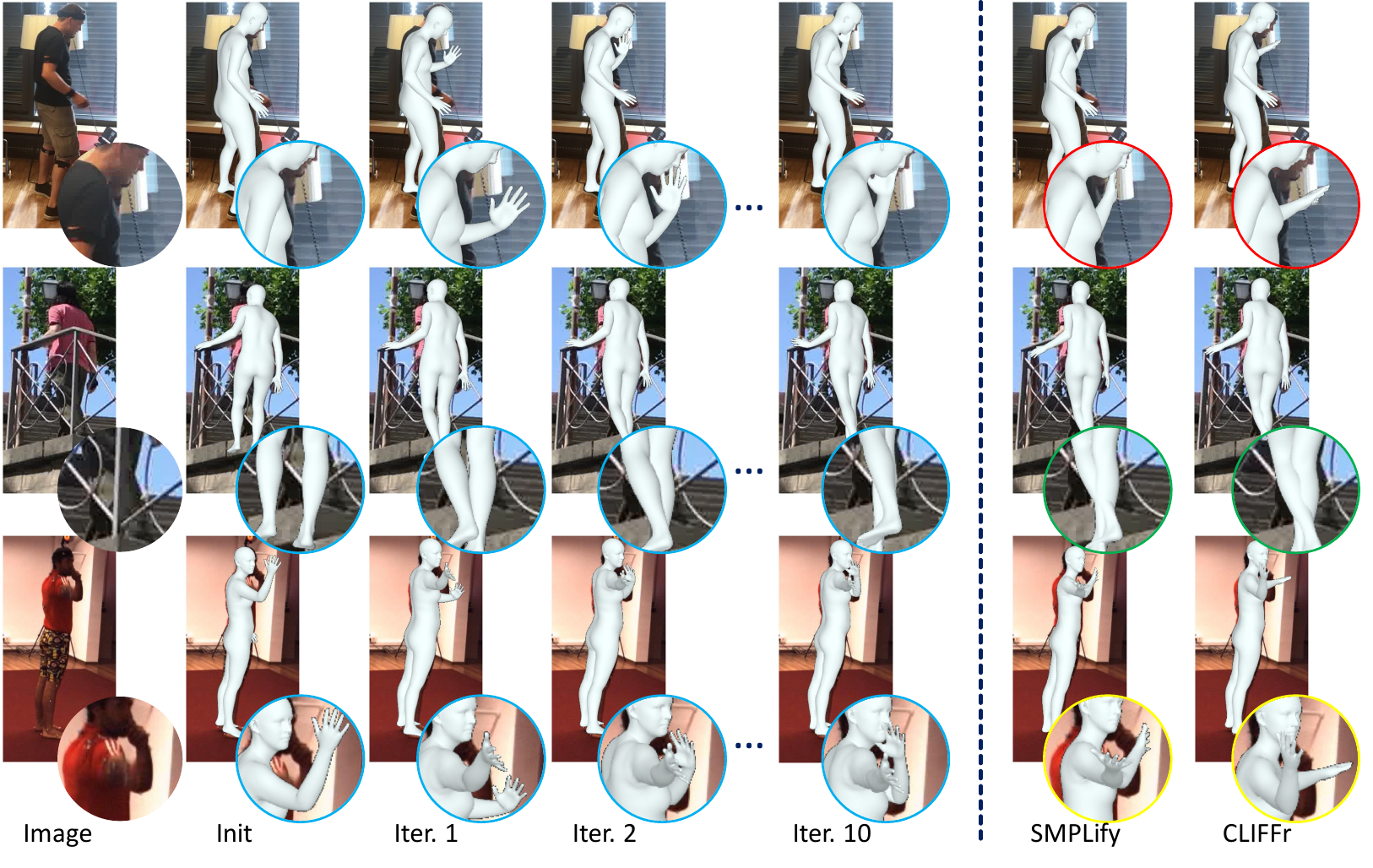}
    \vspace{-0.9cm}
	\caption{Refinement result over iterations (left part) and comparison with other human mesh refinement methods (right part). Left: The blue circles highlight the refinement progress where our method iteratively refines the misaligned bone direction to fit with 2D. Right: the red circles (first row) highlight the refinement of other methods still misaligned with the 2D; the green ones (second row) highlight the penetration problem caused by their depth inaccuracy; and the yellow circles (third row) have both above problems.
 }
	\label{fig:vis_loop}
    \vspace{-0.5cm}
\end{figure*}

\textbf{Iterative Updates.} Fig.~\ref{fig:n_loop} plots the three evaluation metrics with respect to the number of refinement iterations.  %a line chart illustrating performance variations with increasing loop numbers.
The errors gracefully decrease with increasing interactions, converging to the final result after 4-5 iterations. 
Fig.~\ref{fig:vis_loop} shows qualitative examples of body poses over the iterations.  %over the iteillustrates the refinement process for some examples. It's evident that even when 
The initial predictions are misaligned with the visual evidence from the image; our method progressive refines the parameters to match the evidence, while preserving natural poses.

\textbf{Different HMR Models and Representation.} Our method is plug-and-play on top of any HMR method. We test our method on other commonly used base models such as SPIN~\cite{SPIN_2019}, EFT~\cite{EFT_2021} and a different rotation representation PRoM~\cite{gu2023learning}. As shown in Tab.~\ref{tab:otherhmr}, our method can also make large improvements and outperform previous refinement methods with large margins, especially in MPJPE and PVE which are indicators for global orientation and shape.
\begin{table}[t!]
    \centering
    \caption{Results using alternative HMR base models and rotation representations on 3DPW. $^*$ indicates our reproduced results.}
    \vspace{-0.3cm}
    \resizebox{1.0\linewidth}{!}{
    \begin{tabular}{lccc}
        \hline
        Method & PA-MPJPE $\downarrow$ & MPJPE $\downarrow$ & PVE $\downarrow$ \\
        \hline
        SPIN~\cite{SPIN_2019}  & 59.97 & 102.12 & 130.62 \\
        SPIN + SMPLify$^*$ & 47.99 & 87.06 & 102.28 \\
        \rowcolor{lightgray!30}
        SPIN \textbf{+ ours}  & \textbf{42.46} & \textbf{67.12} & \textbf{80.25}  \\
        \hline
        EFTb~\cite{EFT_2021}  & 54.71 & 94.02 & 116.23 \\
        EFTb + SMPLify$^*$ & 44.69 & 82.39 & 96.50 \\
        \rowcolor{lightgray!30}
        EFTb \textbf{+ ours}  & \textbf{32.34} & \textbf{49.14} & \textbf{59.28} \\
        \hline
        PRoM~\cite{gu2023learning}  & 42.0 & 67.6 & 79.2 \\
        PRoM + SMPLify$^*$ & 35.67 & 65.21 & 78.02 \\
        \rowcolor{lightgray!30}
        PRoM \textbf{+ ours}  & \textbf{26.72} & \textbf{42.18} & \textbf{51.25} \\
        \hline
    \end{tabular}}
    \label{tab:otherhmr}
    \vspace{-0.65cm}
\end{table}

%-------------------------------------------------------------------------

\subsection{Comparison with the State-of-the-art}
In this subsection, we compare our approach with the SOTA human mesh refinement methods, as presented in Tab.~\ref{tab:SOTA}.  The upper segment of Tab.~\ref{tab:SOTA} first represents our HMR base model CLIFFb~\cite{CLIFF_2022} as stated previously. The bottom segment ensures a fair comparison as all utilize ground truth 2D keypoints. %Especially, we adapted and concat SMPLify~\cite{SMPLify_2016} to CLIFFb to remain a more fair comparison.
By comparing with optimization-based methods in the second segment (\eg, CLIFFr~\cite{CLIFF_2022} and SMPLify~\cite{SMPLify_2016} adapted by ours),
our KITRO demonstrates a significant enhancement over them (about 15\% improvement in PA-MPJPE, 20\% improvement in MPJPE, 25\% improvement in PVE than SOTA refinement method~\cite{CLIFF_2022}).
Notably, our method achieves a 27.67mm PA-MPJPE joint error on the 3DPW dataset, which is comparably close to the 26mm accuracy obtained in the dataset's creation process~\cite{3DPW_2018}.
This advancement is attributed to our explicit depth and kinematic-tree modeling, which consistently yields a precisely refined pose fitting with 2D clues more ideally.

We also compare our visualization result with other methods as shown in Fig.~\ref{fig:vis_loop}. Based on the same initialized human pose, our approach aligns more accurately with 2D evidence, as evident in the upright position of the left hand in the first row, unlike the incorrect direction seen with other methods. Additionally, while SPMLify~\cite{SMPLify_2016} and CLIFFr~\cite{CLIFF_2022} may correctly fit 2D evidence, depth ambiguities often lead to penetration issues, as in the second row in Fig.~\ref{fig:vis_loop}. Our explicit depth modeling effectively avoids such problems.

%-------------------------------------------------------------------------
\subsection{Analysis}
In this subsection, we analyze KITRO in three aspects: its enhancement on top of other refinement methods; its adaptability to similar parametric models like MANO~\cite{MANO_2017} in hand mesh recovery; and its overall improvement coverage quantified by the improvement samples proportion.

\textbf{Refinement on Top of Refinement.}
Previous subsection shows the result on top of HMR base models (\eg, CLIFFb, SPIN, EFTb). Here we further examine whether our method can get extra improvement on top of other refinement methods. Tab.~\ref{tab:ontopofrefinemment} shows KITRO can further enhance the refinement result from SOTA refinement method CLIFFr, showing KITRO can make gains they are unable to achieve. %In addition, based on CLIFFr leads to a lower refinement than ours on MPJPE and PVE shows that their refinement is .
\begin{table}[t!]
    \centering
    \caption{Our refinement results on top of CLIFFr on 3DPW.}
    \vspace{-0.3cm}
    \resizebox{1.0\linewidth}{!}{
    \begin{tabular}{lccc}
        \hline
        Method & PA-MPJPE $\downarrow$ & MPJPE $\downarrow$ & PVE $\downarrow$ \\
        \hline
        CLIFFb & 43.76 & 73.67 & 91.58 \\
        + CLIFFr  & 32.04 & 55.83 & 71.95 \\
        \rowcolor{lightgray!30}
        + CLIFFr\textbf{ + ours} & \textbf{26.21} & \textbf{46.96} & \textbf{57.53} \\
        \hline
    \end{tabular}}
    \label{tab:ontopofrefinemment}
    \vspace{-0.65cm}
\end{table}

\textbf{Generalizing to Other Parametric Model.}
Beyond the SMPL model, our method's versatility is further illustrated through its application to the MANO model~\cite{MANO_2017}, a commonly used parametric model for hand mesh recovery. Lacking camera prediction in the pre-trained MANO baseline, we utilized ground truth camera translation for these experiments and ensured the same setting for other adapted refinement methods for fair comparison. The results shown in Tab.~\ref{tab:mano} demonstrate that our method not only refines human mesh predictions effectively but is also suitable for other parametric models like MANO.

\begin{table}[t!]
    \centering
    \caption{Result of our method adopted to MANO model on Freihand. PA-PVE denotes the PVE result after Procrustes aligned.}
    \vspace{-0.3cm}
    \resizebox{0.95\linewidth}{!}{
    \begin{tabular}{lccc}
        \hline
        Method & PA-MPJPE $\downarrow$ & PA-PVE $\downarrow$ \\
        \hline
        Baseline (MANO)*  & 7.82 & 8.01 \\
        Baseline + SMPLify* & 4.66 & 5.07 \\
        \rowcolor{lightgray!30}
        Baseline\textbf{ + ours} & \textbf{4.06} & \textbf{4.58} \\
        \hline
    \end{tabular}}
    \label{tab:mano}
    \vspace{-0.65cm}
\end{table}

\textbf{Comprehensive Improvement Across Joints and Samples.} Our method not only enhances average performance but also ensures equitable and widespread effectiveness across joints and samples. As depicted in \cref{fig:teaser:improv_joints}, it effectively enhances both proximal and distal joints, while other methods may sacrifice the proximal ones. Additionally, a significant portion of individual samples improved after our refinement: 88\% on 3DPW and 92\% on Human3.6M as shown in Sec. \textcolor{red}{E} of the Supplementary. This underscores our method's comprehensive efficacy.

\section{Limitations}
KITRO relies on the initial predicted mesh as a reference for hypothesis selection, which can lead to deviations in cases of poor initial mesh predictions. This is particularly evident in scenarios of person misidentification errors~\footnote{Predicting the wrong person in an image with multiple individuals.}, where the initial 3D mesh is totally mismatched with the 2D evidence. More discussion is in Sec. \textcolor{red}{F} of the Supplementary.
In addition, similar to other human mesh refinement methods, KITRO relies on the accuracy of the input 2D pose. A detailed discussion and evaluation on the detected and noisy 2D keypoints can be found in Sec. \textcolor{red}{G} of the Supplementary.

\section{Conclusion}
Motivated by the inadequate depth accuracy and suboptimal proximal joint performance in existing human mesh refinement methods, we propose KITRO. By explicitly modeling joint depth in closed-form solution along the human kinematic-tree, we can then use the decision tree to trace all hypotheses and select the most likely one. KITRO demonstrates superior accuracy and comprehensive improvement across various datasets and baseline models.

\section*{Acknowledgement}
This research / project is supported by the Ministry of Education, Singapore, under its MOE Academic Research Fund Tier 2 (STEM RIE2025 MOE-T2EP20220-0015).
{
    \small
    \bibliographystyle{ieeenat_fullname}
    \bibliography{main}
}

\clearpage
\setcounter{page}{1}
% \title{KITRO: Refining Human Mesh by 2D Clues and Kinematic-tree Rotation \\ Supplementary Material}
\maketitlesupplementary

% Reset the section counter
\setcounter{section}{0}
% Redefine the section numbering to use letters
\renewcommand{\thesection}{\Alph{section}}

\section{Details on Swing-Twist Decomposition} \label{supp:swingtwist}
\begin{figure}[ht]
    \vspace{-0.2cm}
	\centering
	\includegraphics[width=1.0\linewidth]{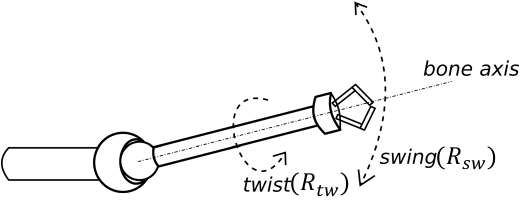}
    \vspace{-0.5cm}
	\caption{Illustration of the swing-twist decomposition~\cite{SwingTwist_2015}.}
	\label{fig:swing_twist}
    \vspace{-0.3cm}
\end{figure}

Swing-twist decomposition for a rotation is a fundamental technique in computer graphics~\cite{SwingTwist_1998,SwingTwist_2001,SwingTwist_2015} and was first introduced to the human mesh recovery field by HybrIK~\cite{Hybrik_2021}. As shown in Fig.~\ref{fig:swing_twist}, for any given joint, its rotation \( \boldsymbol{\theta}_R \in \mathbb{SO}(3) \) can be decomposed into a 2 degree-of-freedom (DoF) swing rotation \( R_{sw} \) and a 1 DoF twist rotation \( R_{tw} \), such that \( \boldsymbol{\theta}_R = R_{sw} R_{tw} \).

\textbf{Swing Rotation.} $R_{s w}$ is to rotate the bone from the template relative pose $\vec{t_r}$ to the target relative orientation $\vec{p_r}$. This rotation occurs around axis \( \vec{n} \), defined as:
\begin{equation}
\vec{n}=\frac{\vec{t_r} \times \vec{p_r}}{\|\vec{t_r} \times \vec{p_r}\|},
\end{equation}
which is orthogonal to both \( \vec{t_r} \) and \( \vec{p_r} \). The rotation magnitude, denoted as \( \gamma \), is the angle subtended between \( \vec{t_r} \) and \( \vec{p_r} \):
\begin{equation}
\cos \gamma=\frac{\vec{t_r} \cdot \vec{p_r}}{\|\vec{t_r}\|\|\vec{p_r}\|}, \quad \sin \gamma=\frac{\|\vec{t_r} \times \vec{p_r}\|}{\|\vec{t_r}\|\|\vec{p_r}\|}
\end{equation}
Employing Rodrigues' rotation formula, the swing rotation \( R_{sw} \) is expressed in closed form as:
\begin{equation}
R_{s w}=\mathcal{I}+\sin \gamma[\vec{n}]_{\times}+(1-\cos \gamma)[\vec{n}]_{\times}^2,
\end{equation}
where \( \mathcal{I} \) represents the \( 3\times3 \) identity matrix and \( [\vec{n}]_{\times} \) denotes the skew-symmetric matrix of \( \vec{n} \).

\textbf{Twist Rotation.} Then $R_{t w}$ is to rotate the bone around bone axis $\vec{t_r}$ itself. Let \( \varphi \) represent the rotation angle. According to Rodrigues' formula, the twist rotation is given by:
\begin{equation}
R_{t w}=\mathcal{I}+\sin \varphi \frac{[\vec{t_r}]_{\times}}{||\vec{t_r}||}+(1-\cos \varphi) \frac{[\vec{t_r}]_{\times}^2}{||\vec{t_r}||^2},
\end{equation}
where \( [\vec{t_r}]_{\times} \) is the skew-symmetric matrix of \( \vec{t_r} \).

In our work, we focus on refining the swing rotation \( R_{sw} \) while preserving the initial twist rotation \( R_{tw} \) estimates. This is motivated by the limited variability in twist angles $\varphi$ due to human physiological constraints, as evidenced by HybrIK's empirical studies~\cite{Hybrik_2021}. In contrast, the swing rotation \( R_{sw} \) exhibits a more significant range of motion, necessitating a more detailed and accurate refinement. Thus, in our approach, we explicitly formulate the bone directions in closed form in order to refine the swing rotation \( R_{sw} \).

\section{Proof-of-concept for Solution Selection} \label{supp:proofofconcept}
\begin{figure}[h]
    \vspace{-0.2cm}
	\centering
	\includegraphics[width=1.0\linewidth]{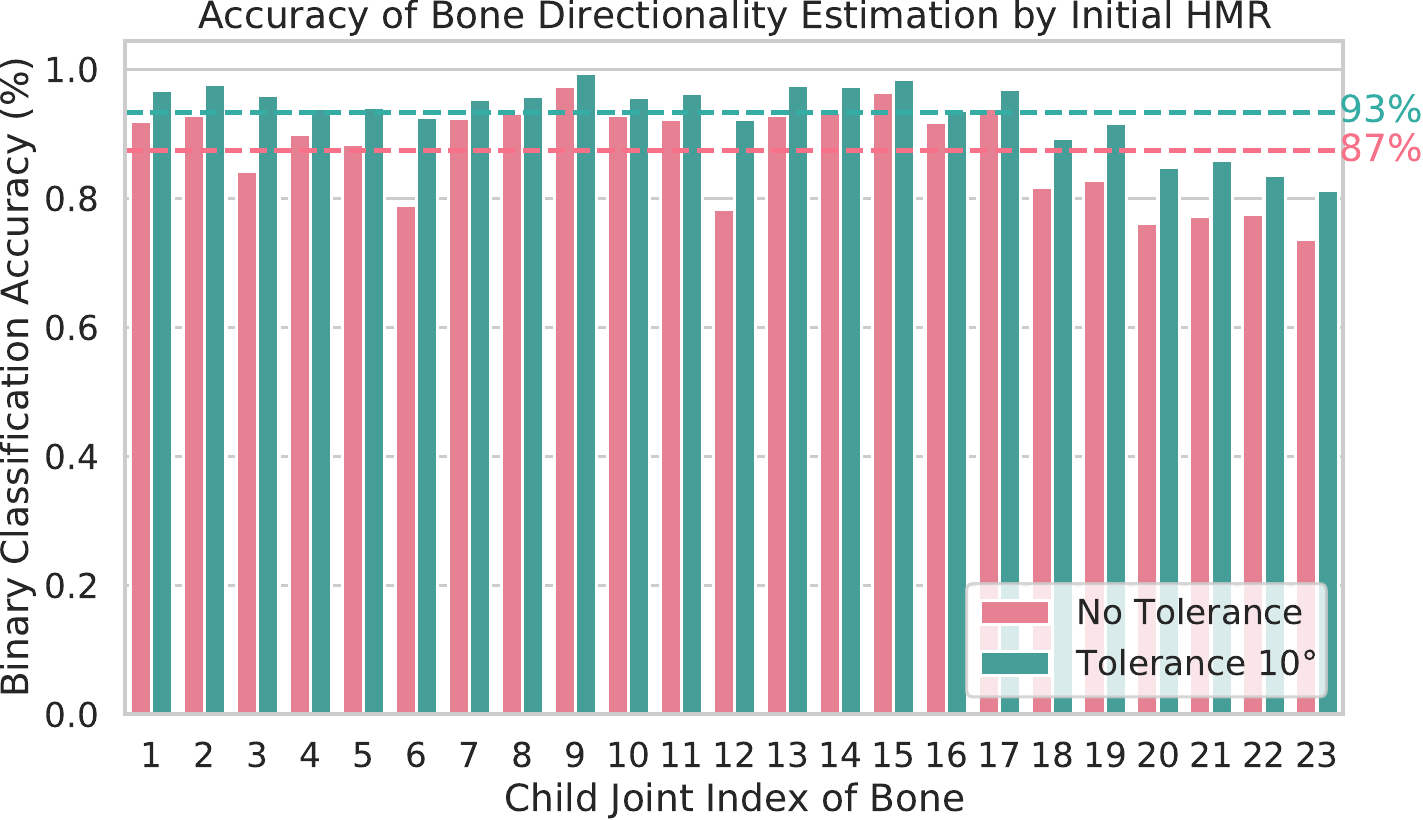}
    \vspace{-0.6cm}
	\caption{Classification accuracy of bone pointing towards or away from the camera using the HMR model. Red bars indicate 87\% accuracy without tolerance, while green bars show improved accuracy up to 93\% when a $10^{\circ}$ error margin is allowed.  These results highlight the HMR model's effectiveness in coarse-grained directionality estimation of bones.
    }
	\label{fig:towardaway}
    \vspace{-0.4cm}
\end{figure}

In this section, we conduct empirical studies to validate the assumptions discussed in Sec. \textcolor{red}{4.3} of the main paper. As shown by the red bars in Fig.~\ref{fig:towardaway}, 87\% of bones in the initial HMR estimates are correctly identified as pointing towards or away from the camera. When a $10^{\circ}$ margin of error is tolerated in ambiguous cases where only 10 degrees separate two solutions, the accuracy increases to 93\% as illustrated by the green bars in Fig.~\ref{fig:towardaway}. These results affirm the effectiveness of the original HMR model in determining bone direction towards or away from the camera, providing a reliable prior for the decision tree formulation of our method.

\section{Correctness Proof for $\boldsymbol{\theta}$ update } \label{supp:proof}
In this section, we validate the correctness of the pose update equation (\ie, Eq. \textcolor{red}{19} in Sec. \textcolor{red}{4.3} of the main paper).
\begin{proof}
Assuming the refinement from the root joint up to the parent joint $\Tilde{p}$ of joint $p$ is already done, the current focus is updating rotation ${\boldsymbol{\theta}_R^p}$ for joint $p$. The objective is to verify that updating the joint rotation from $\boldsymbol{\theta}_R^p$ to ${\boldsymbol{\theta}_R^p}'$ correctly rotate the bone direction of bone $(p,c)$ from $\vec{\mathbf{b}}^{(p,c)}$ to $\vec{\mathbf{b}}^{(p,c)}_{new}$. Considering the joint rotations in $\boldsymbol{\theta}$ are all relative to each parent joint's coordinate system, the bone direction in the absolute coordinate system is derived as the product of relative rotations from the root joint to the parent joint along the kinematic chain. Hence, the global rotation for joint $p$ in the absolute coordinate system before refinement is:
\begin{equation} \label{eq:relative2absolute}
    R_{abs}^{p} = \prod_{i \in KC(p)} \boldsymbol{\theta}_R^i,
\end{equation}
where $\prod_{i \in KC(p)}$ denotes the matrix product of rotation matrices from the root joint to joint $p$, with $KC(p)$ representing the kinematic chain.  Considering the template relative pose (T pose) for bone $(p,c)$ denoted as $\vec{t}_r^p$, when applied with absolute rotation from Eq.~\ref{eq:relative2absolute}, yields the absolute bone direction:
\begin{equation} \label{eq:relative2absolutedir}
    \vec{\mathbf{b}}^{(p,c)} = R_{abs}^{p} \cdot \vec{t}^p_r = \prod_{i \in KC(p)} \boldsymbol{\theta}_R^i \cdot \vec{t}^p_r,
\end{equation}
As mentioned in the main paper, $R_{sw}^{(p,c)}$, computed via Rodrigues' formula, is the rotation matrix that rotates $\vec{\mathbf{b}}^{(p,c)}$ to $\vec{\mathbf{b}}^{(p,c)}_{new}$ in the absolute coordinate system:
\begin{equation} \label{eq:rotb2new}
    \vec{\mathbf{b}}^{(p,c)}_{new} = R_{sw}^{(p,c)} \cdot \vec{\mathbf{b}}^{(p,c)}.
\end{equation}
Now we verify the updated absolute rotation for joint $p$ after the refinement of Eq. \textcolor{red}{19}:
\begin{equation} \label{eq:absolutebonedir}
    {R_{abs}^{p}}' = \prod_{i \in KC(\Tilde{p})} \boldsymbol{\theta}_R^i \cdot {\boldsymbol{\theta}_R^p}',
\end{equation}
applying this new rotation to the T pose for bone $(p,c)$ results in:
\begin{equation} \label{eq:proof}
\begin{aligned}
    {R_{abs}^{p}}' \cdot \vec{t}^p_r =& \prod_{i \in KC(\Tilde{p})} \boldsymbol{\theta}_R^i \cdot {\boldsymbol{\theta}_R^p}' \cdot \vec{t}^p_r && \text{\small{(from Eq.~\ref{eq:absolutebonedir})}}\\
    =& \prod_{i \in KC(\Tilde{p})} \boldsymbol{\theta}_R^i \cdot (\prod_{i \in KC(\Tilde{p})} \boldsymbol{\theta}_R^i)^T \\
    &\cdot R_{sw}^{(p,c)} \cdot \prod_{i \in KC(p)} \boldsymbol{\theta}_R^i \cdot \vec{t}^p_r  && \text{\small{(Main's Eq. \textcolor{red}{19})}} \\
    =& R_{sw}^{(p,c)} \cdot \prod_{i \in KC(p)} \boldsymbol{\theta}_R^i \cdot \vec{t}^p_r && \text{\small{(SymMat property)}}\\
    =& R_{sw}^{(p,c)} \cdot \vec{\mathbf{b}}^{(p,c)} && \text{\small{(from Eq.~\ref{eq:relative2absolutedir})}}\\
    =& \vec{\mathbf{b}}^{(p,c)}_{new} && \text{\small{(from Eq.~\ref{eq:rotb2new})}}\\
\end{aligned}
\end{equation}
The `SymMat property' corresponds to the property of symmetry rotation matrices, where the transpose of a rotation matrix equals its inverse. Eq.~\ref{eq:proof} demonstrates that the updated joint rotation ${\boldsymbol{\theta}_R^p}'$ correctly modifies $\vec{\mathbf{b}}^{(p,c)}$ to the desired absolute rotation $\vec{\mathbf{b}}^{(p,c)}_{new}$.
\end{proof}

\section{More Ablation Studies} \label{supp:moreablation}
In this section, we present extended ablation studies for our framework design. 

Tab.~\ref{tab:moreablation3DPW} details the results of eight different configurations of camera, shape, and pose refinement. The study reveals that when refining only one of these three factors fails to achieve effective results. This ineffectiveness can be attributed to the importance of 2D keypoints and 3D human mesh alignment in our approach. Without the proposed alignment mechanisms in camera and shape refinements, pose refinements alone are also insufficient.

\begin{table}[h!]
    \vspace{-0.3cm}
    \centering
    \caption{Detailed ablation study for three factors on 3DPW dataset. The first row represents the baseline HMR model, and the last row depicts our full model. The identical results in the first two rows are because updating only the camera does not change SMPL parameters, leading to unchanged outcomes.}
    \vspace{-0.2cm}
    \resizebox{1.0\linewidth}{!}{
    \begin{tabular}{cccccc}
        \hline
        Camera & Shape & Pose & PA-MPJPE $\downarrow$ & MPJPE $\downarrow$ & PVE $\downarrow$ \\
        \hline 
        \ding{55} & \ding{55} & \ding{55} & 43.76 & 73.67 & 91.58 \\
        \checkmark  & \ding{55} & \ding{55} & 43.76 & 73.67 & 91.58\\
        \ding{55} & \checkmark & \ding{55}  & 44.31 & 69.92 & 83.26\\        
        \ding{55} & \ding{55} &  \checkmark & 45.92 & 87.33 & 100.73 \\
        \checkmark & \checkmark  & \ding{55} & 44.57 & 80.03 & 95.71\\
        \checkmark & \ding{55} & \checkmark & 35.00 & 69.02 & 84.25\\
        \ding{55} & \checkmark & \checkmark  & 28.53 & 46.68 & 57.76 \\
        \checkmark & \checkmark & \checkmark & \textbf{27.67} & \textbf{43.53} & \textbf{53.44} \\
        \hline
    \end{tabular}}
    \label{tab:moreablation3DPW}
    \vspace{-0.3cm}
\end{table}

\begin{figure}[h]
    \vspace{-0.2cm}
	\centering
	\includegraphics[width=1.0\linewidth]{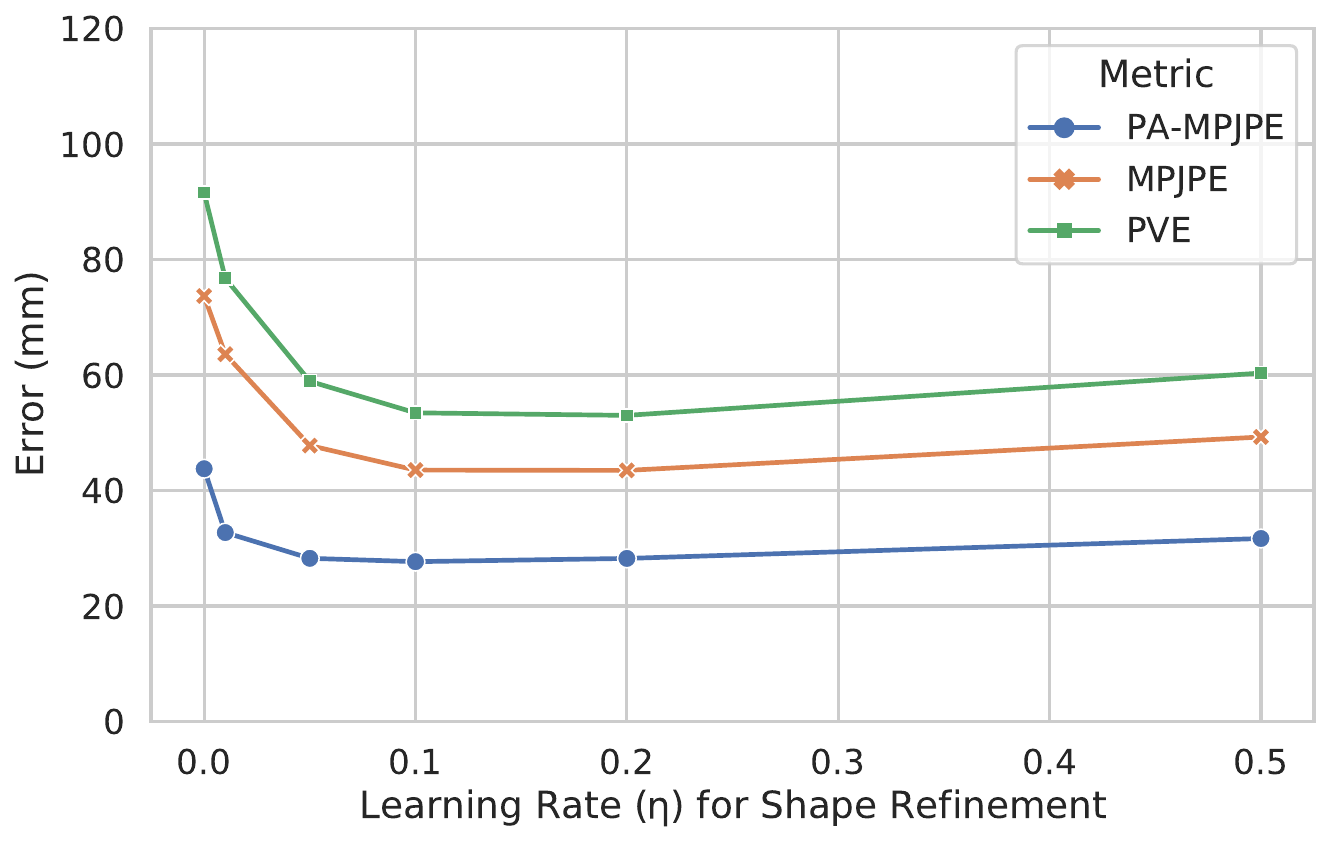}
    \vspace{-0.6cm}
	\caption{Error with different learning rates for shape refinement. These three line plots (PA-MPJPE, MPJPE, and PVE) illustrate our model's performance stability across a range of learning rates.}
	\label{fig:shapelr}
    \vspace{-0.4cm}
\end{figure}

\begin{figure}[h]
    \vspace{-0.2cm}
	\centering
	\includegraphics[width=1.0\linewidth]{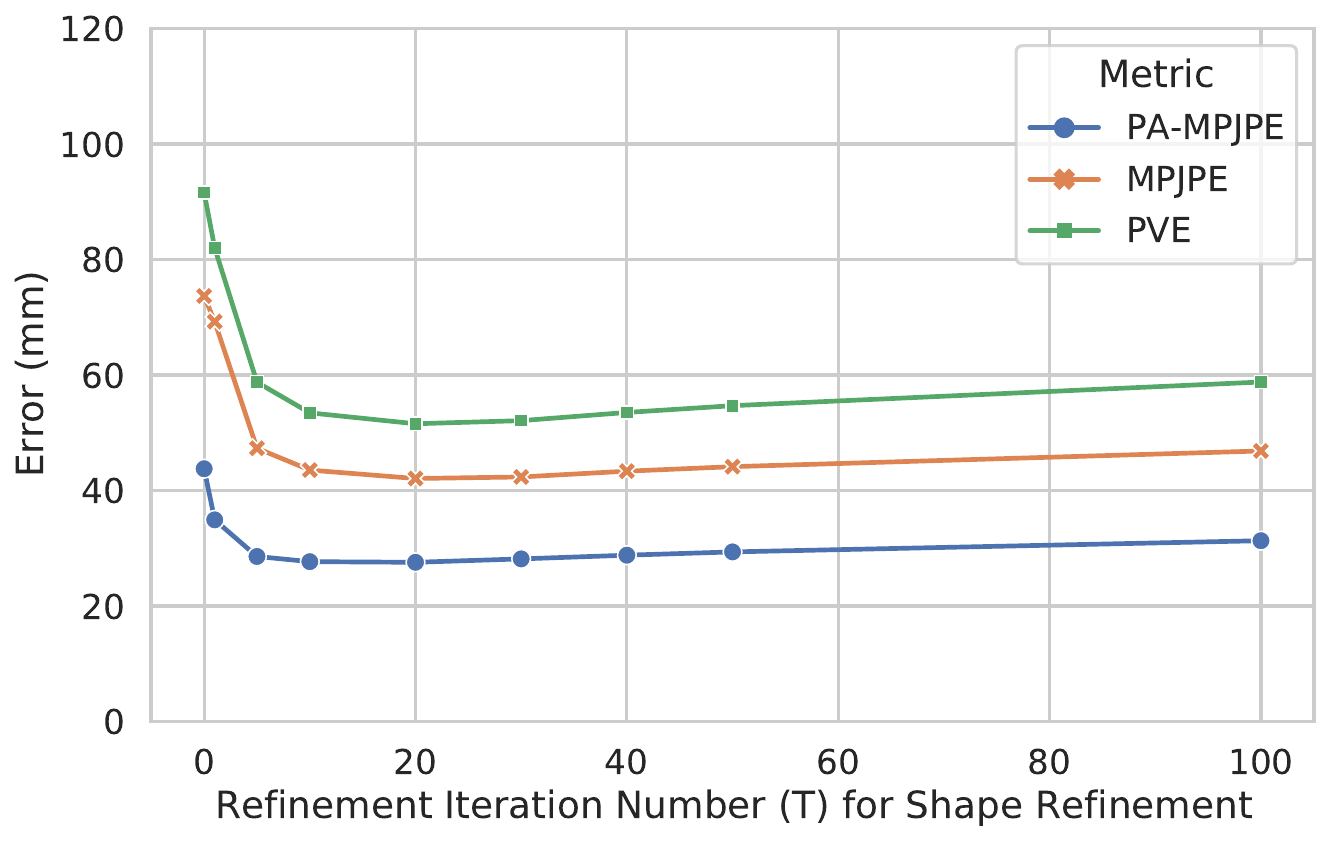}
    \vspace{-0.6cm}
	\caption{Error with the number of fine-tuning iterations. The results indicate consistent model performance over varying iterations for shape refinement.}
	\label{fig:shapeiter}
    \vspace{-0.4cm}
\end{figure}

In addition, we explored varying learning rates (Fig.~\ref{fig:shapelr}) and refinement iteration numbers (Fig.~\ref{fig:shapeiter}) for the shape refinement model as discussed in Sec \textcolor{red}{5.2} of the main paper. The results depicted in these figures demonstrate that both the learning rate and iteration number do not significantly impact the performance of our method. This indicates a robustness in our approach to variations in these parameters.

\section{Improvement Distribution over Samples} \label{supp:Distribution}
\begin{figure}[h]
    \vspace{-0.2cm}
	\centering
	\includegraphics[width=1.0\linewidth]{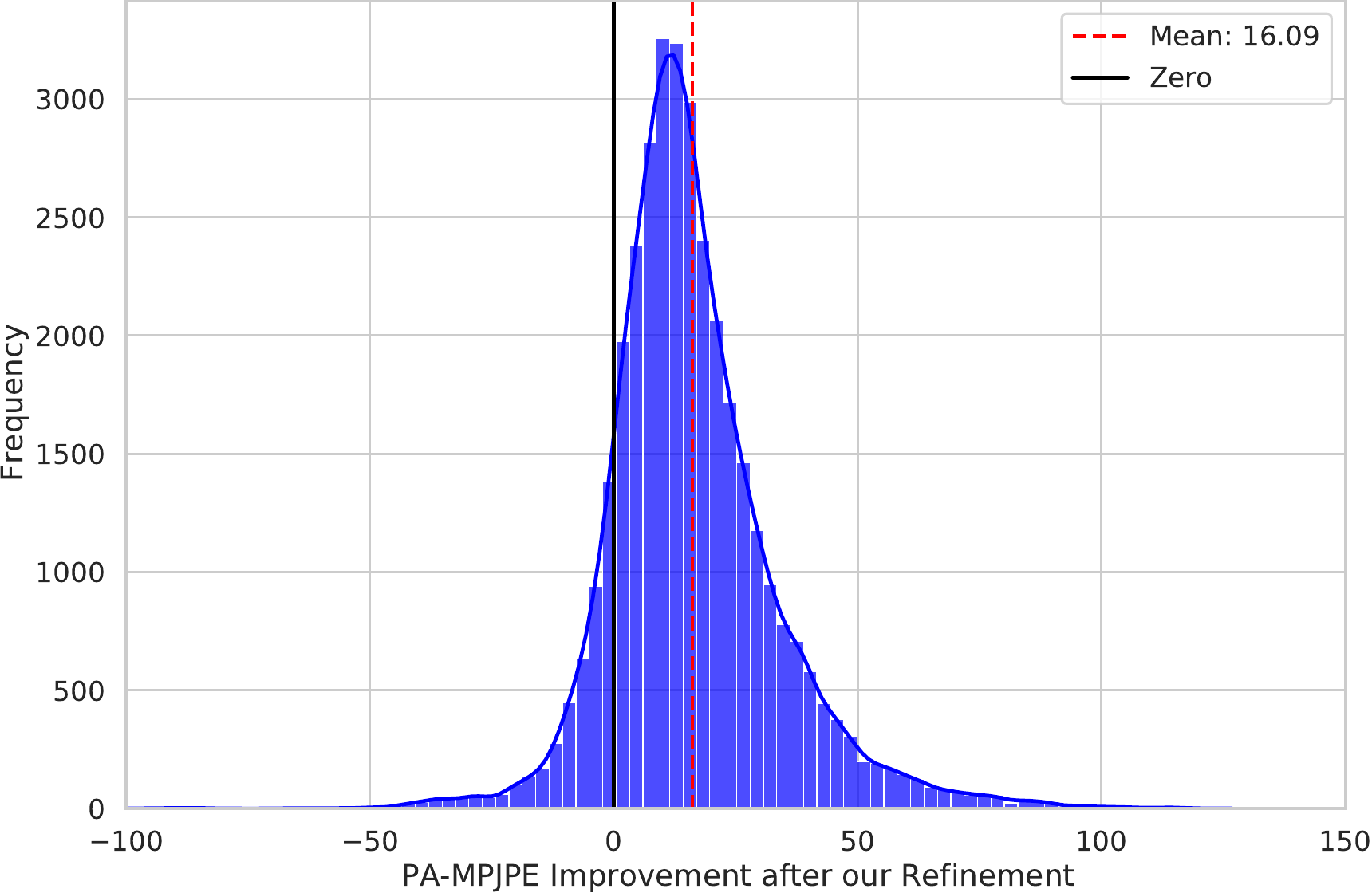}
    \vspace{-0.6cm}
	\caption{Distribution of performance improvement on 3DPW.}
	\label{fig:3DPW_improvement}
    \vspace{-0.4cm}
\end{figure}

\begin{figure}[h]
    \vspace{-0.2cm}
	\centering
	\includegraphics[width=1.0\linewidth]{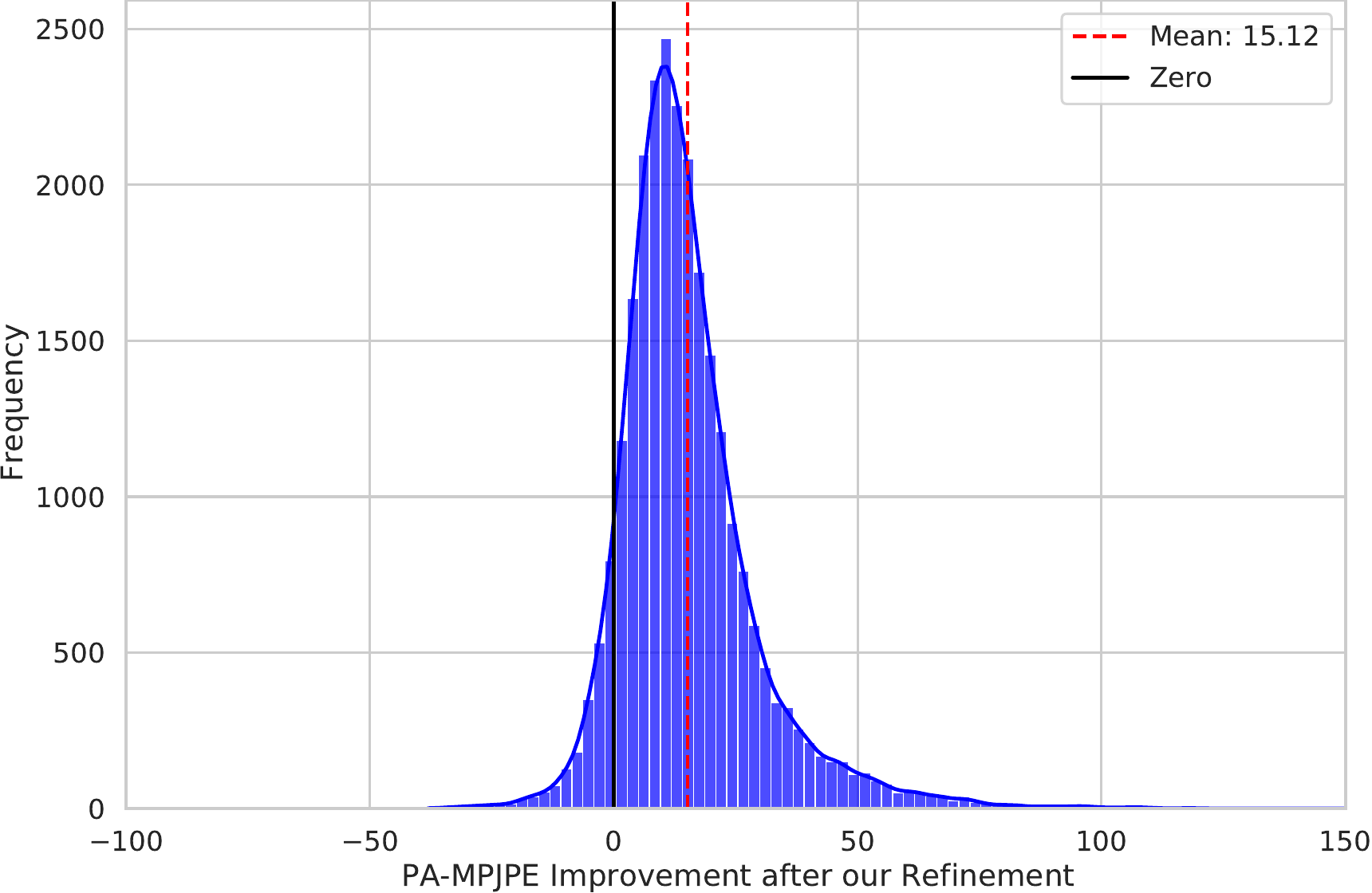}
    \vspace{-0.6cm}
	\caption{Distribution of performance improvement on the Human3.6M.}
	\label{fig:Human36M_improvement}
    \vspace{-0.4cm}
\end{figure}

In this section, we illustrate the extent of improvement in individual samples following our refinement process. Fig.~\ref{fig:3DPW_improvement} and Fig.~\ref{fig:Human36M_improvement} displays the distribution of performance improvements on the 3DPW and Human3.6M dataset respectively. These visualizations demonstrate that a majority of the samples exhibit significant improvement, again demonstrating the comprehensive efficacy of our method.

\begin{figure}[h]
    \vspace{-0.4cm}
    \centering
    \begin{subfigure}[b]{0.235\textwidth}
        \centering
        \includegraphics[width=\textwidth]{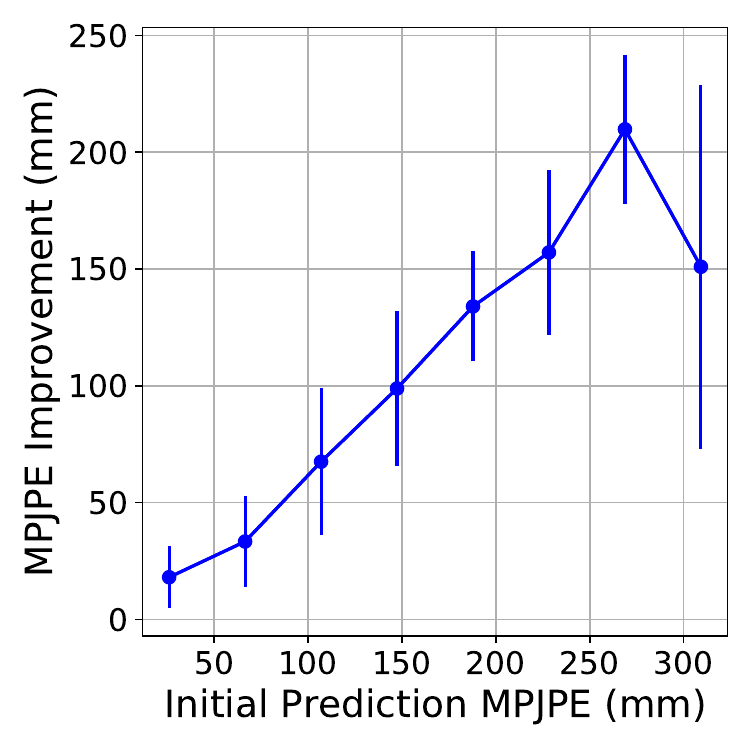}
        \vspace{-0.5cm}
        \caption{Our robustness to high-errors.}
        \label{fig:improvement_initerr}
    \end{subfigure}
    \begin{subfigure}[b]{0.235\textwidth}
        \centering
        \includegraphics[width=\textwidth]{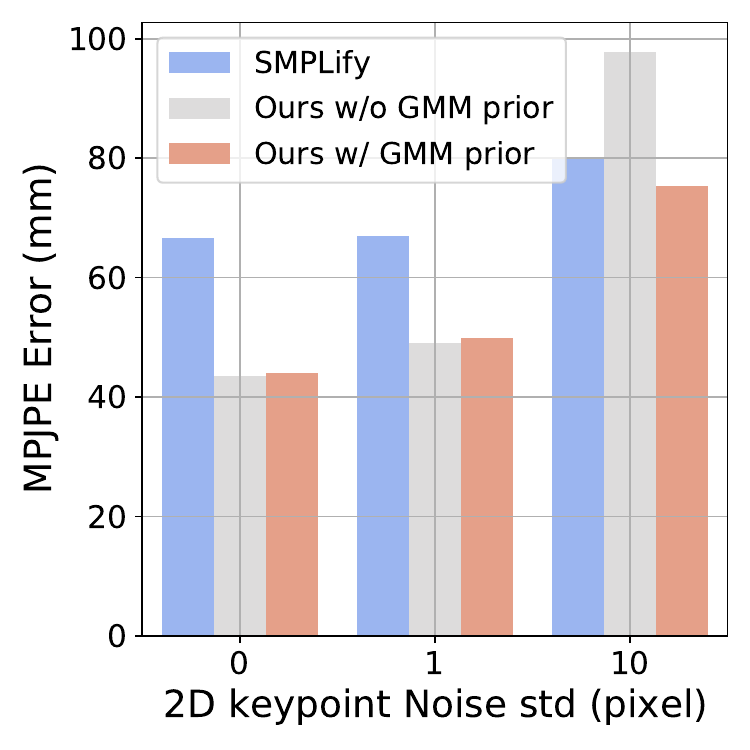}
        \vspace{-0.5cm}
        \caption{Impact of 2D keypoint quality.} 
        \label{fig:2dnoise}
    \end{subfigure}
    \vspace{-0.25cm}
    \label{fig:supp}
    \vspace{-0.6cm}
\end{figure}

\begin{figure}[b]
    \vspace{-0.4cm}
	\centering
	\includegraphics[width=1.0\linewidth]{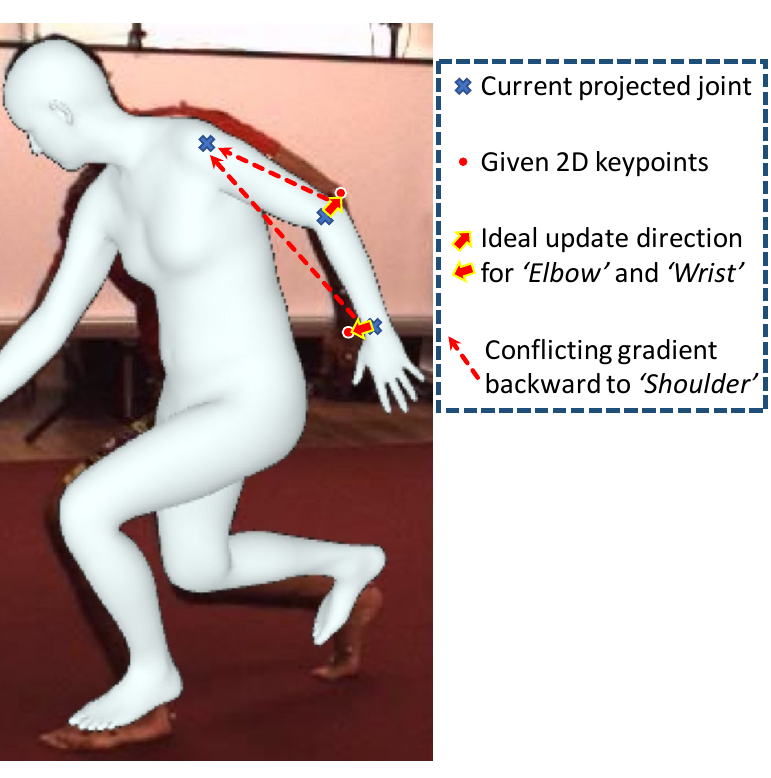}
    \vspace{-0.6cm}
	\caption{Illustration of gradient conflict: conflicting gradients at the Elbow and Wrist can complicate and negatively impact the optimization of the Shoulder joint, demonstrating a key limitation in previous gradient-based human mesh refinement methods.}
	\label{fig:conflictgredient}
    \vspace{-0.4cm}
\end{figure}

\section{Impact of Initial Predicted Mesh Quality}
As mentioned in the limitation section, our method relies on the initial predicted mesh as a reference for hypothesis selection. Here we investigate how poor initial mesh predictions could impact the improvement. As shown in Fig.~\ref{fig:improvement_initerr}, where we plot the MPJPE improvement \wrt the original MPJPE error, KITRO withstands high errors with consistently larger improvements. The reason is our global path selection with the decision tree tolerates a few initial bone-facing errors. Additionally, about half of the 13\% bone-facing errors are from similar-valued solutions which are not serious mistakes. However, when the initial prediction is too wrong, e.g. 300mm MPJPE or too many bone-facing errors (see Tab.~\ref{tab:bonefacingerr}), then our improvements decrease.

\begin{table}[h!]
    \vspace{-0.3cm}
    \centering
    \caption{Impact of base models on bone-facing and PA-MPJPE.}
    \vspace{-0.3cm}
    \resizebox{0.85\linewidth}{!}{
    \begin{tabular}{lccc}
    \hline
    \multicolumn{1}{c}{\multirow{2}{*}{Base Model}} & \multicolumn{1}{c}{{\# of Correct Bone-facing}} & \multicolumn{2}{c}{PA-MPJPE}                           \\ \cline{3-4} 
    \multicolumn{1}{c}{}                            & \multicolumn{1}{c}{(out of 23 bones)}                                        & \multicolumn{1}{c}{Ours} & \multicolumn{1}{c}{SMPLify} \\ \hline
    CLIFFb & 20.1 $\pm$ 1.8 & 27.67 & 36.11 \\
    EFT    & 19.5 $\pm$ 2.1 & 32.34 & 44.69 \\
    SPIN   & 18.8 $\pm$ 2.4 & 42.46 & 47.99 \\ \hline
    \end{tabular}}
    \label{tab:bonefacingerr}
    \vspace{-0.3cm}
\end{table}

\section{Impact of Input 2D Keypoint Quality}
We follow the protocol of previous human mesh refinement works~\cite{CLIFF_2022,EFT_2021, pose2mesh_2020, SMPLify_2016, DynaBOA_2022}, we and all these works refine 3D pose and shape estimates with ground truth 2D keypoints. Given the different conventions between 2D detectors and SMPL regarding the definition of joints~\cite{SMPL_2015}, it's non-trivial to directly use detected 2D keypoints. For example, SMPL puts the hip joint where the bone rotation happens while Openpose~\cite{OpenPose_2017} locates it at the surface landmark where the thigh begins, so one needs to define or learn a mapping from Openpose to SMPL format, or directly train a detector upon SMPL format. Here we performed a rudimentary experiment to train a basic neural network architecture---a simple 3-layer MLP---on 3DPW training data to get the mapping from Openpose to SMPL format. As shown in Tab.~\ref{tab:detectedKPresult}, the performance of both our method and SMPLify diminishes when subjected to noisy 2D keypoints detection and mapping. This degradation is expected as human mesh refinement inherently relies on the precision of 2D keypoints. Nevertheless, our method is still more robust than SMPLify under the detected keypoints scenario. It is important to note that the current approach, utilizing only an MLP for keypoint format mapping, leaves room for enhancement through more advanced mapping strategies and more training data. However, those explorations extend beyond the scope of this paper and are left for future investigation.

\begin{table}[t!]
    \centering
    \caption{Refinement results using 2D keypoints mapping from Openpose detection result. `2DKP' denotes '2D Keypoints'; `GT' denotes 'Ground Truth'. }
    \vspace{-0.2cm}
    \resizebox{1.0\linewidth}{!}{
    \begin{tabular}{lcccc}
        \hline
        Method & 2DKP & PA-MPJPE $\downarrow$ & MPJPE $\downarrow$ & PVE $\downarrow$ \\
        \hline
        SMPLify & GT & 39.99 & 71.11 & 84.28 \\
        \textbf{Ours} & GT & 27.45 & 48.42 & 59.65 \\
        % \rowcolor{lightgray!30}
        % \textbf{Ours} & GT & \textbf{42.46} & \textbf{67.12} & \textbf{80.25}  \\
        \hline
        SMPLify & Detected & 51.35 & 90.68 & 107.41 \\
        % \textbf{Ours (w/o GMM)} & Detected & 50.06 & 83.56 & 99.94 \\
        \textbf{Ours} & Detected & \textbf{45.88} & \textbf{79.80} & \textbf{96.60} \\
        \hline
    \end{tabular}}
    \label{tab:detectedKPresult}
    \vspace{-0.5cm}
\end{table}

Furthermore, we add Gaussian noise under different standard deviations to simulate the poor 2D keypoint quality as shown in Fig.~\ref{fig:2dnoise}.
Our raw method inherently assumes that the 2D pose is correct and as such, can only withstand small errors (see Fig.~\ref{fig:2dnoise}, 1-2 px std). This assumption does not hold for larger errors and some intervention is required. A simple strategy is to add a GMM prior, similar to SMPLify.  SMPLify uses the GMM as a loss;  we use it as a likelihood to reject super unnaturally refined outputs based on erroneous 2D poses (see Fig.~\ref{fig:2dnoise}, 10 px std). More effective filtering strategies to improve overall robustness to 2D pose error are outside our current scope and are left for future work.

\section{Joints and Bones Names on Kinematic-tree} \label{supp:nameKT}
\begin{figure*}[t]
    \vspace{-0.2cm}
	\centering
	\includegraphics[width=0.9\linewidth]{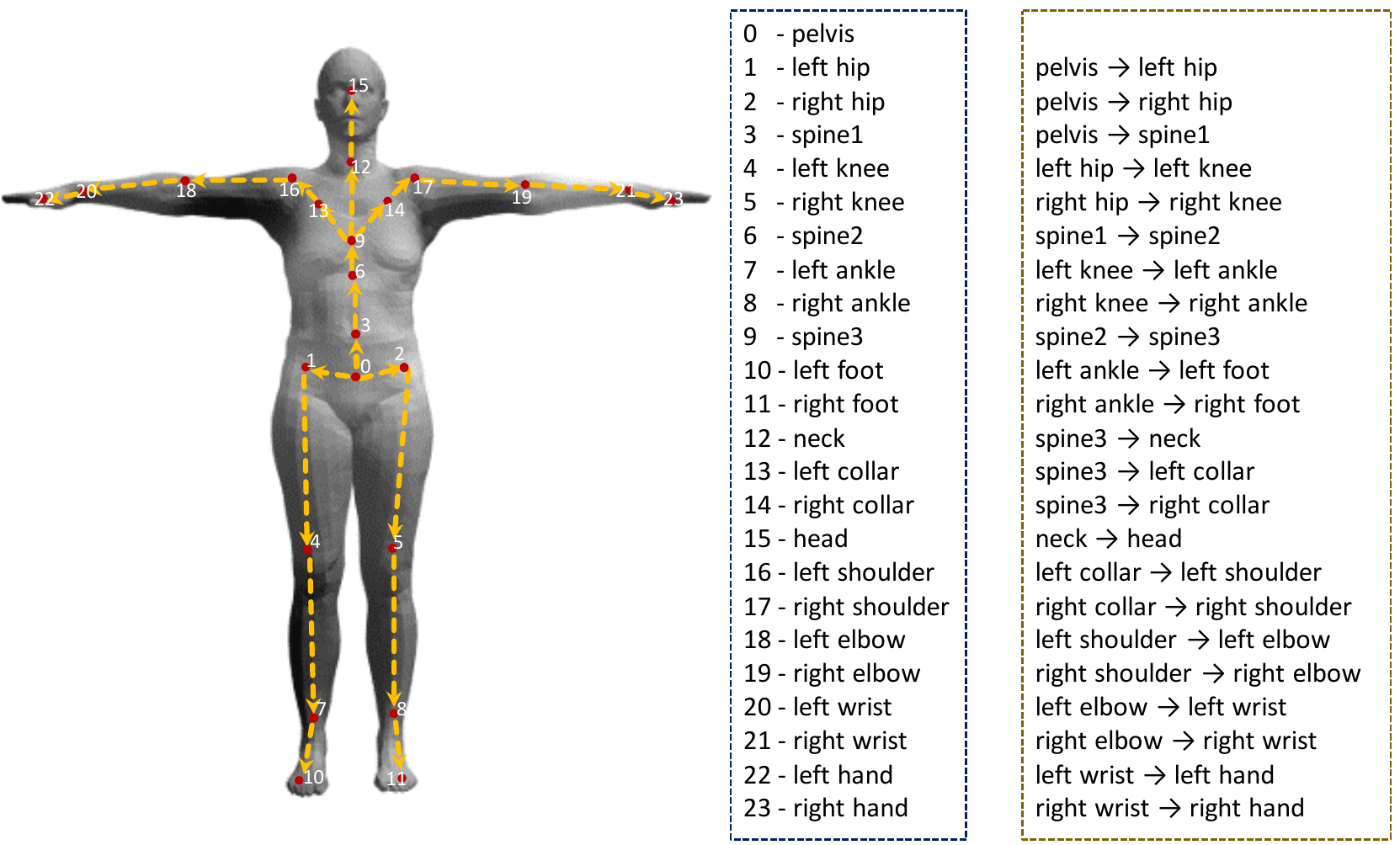}
    \vspace{-0.3cm}
	\caption{The definition of all 24 joints and 23 bones in the human kinematic-tree used by our approach.}
	\label{fig:jointsnames}
    \vspace{-0.2cm}
\end{figure*}

According to SMPL~\cite{SMPL_2015}, there are 24 joints and 23 bones defined in the human kinematic-tree. We list all joint names and bones in Fig.~\ref{fig:jointsnames} 

\section{More Visualization Results}
\begin{figure*}[t]
    % \vspace{-0.1cm}
	\centering
	\includegraphics[width=1.0\linewidth]{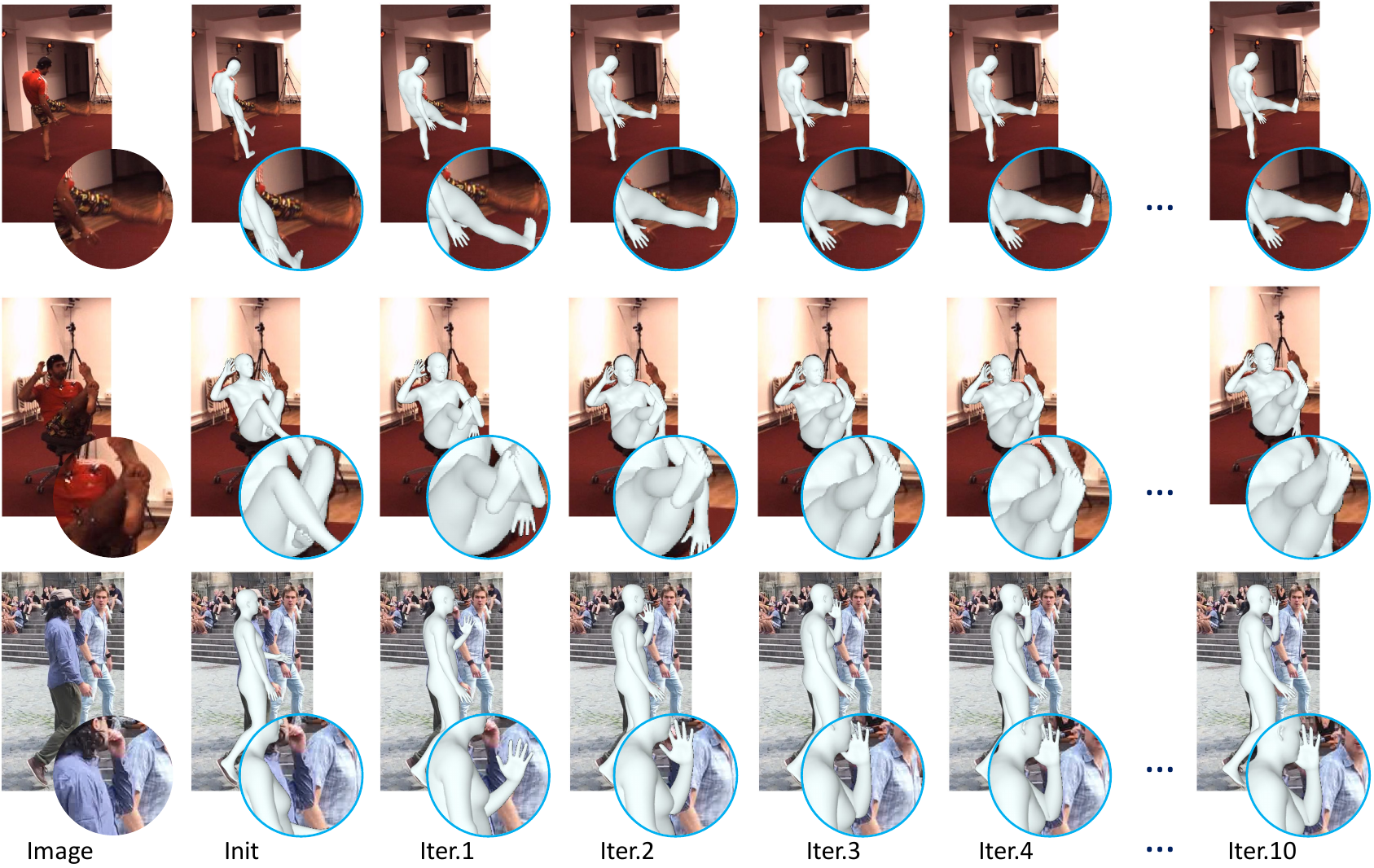}
    \vspace{-0.6cm}
	\caption{More iterative refinement visualizations.}
	\label{fig:moreloops}
    \vspace{-0.2cm}
\end{figure*}

\begin{figure*}[t]
    \vspace{-0.65cm}
	\centering
	\includegraphics[width=0.9\linewidth]{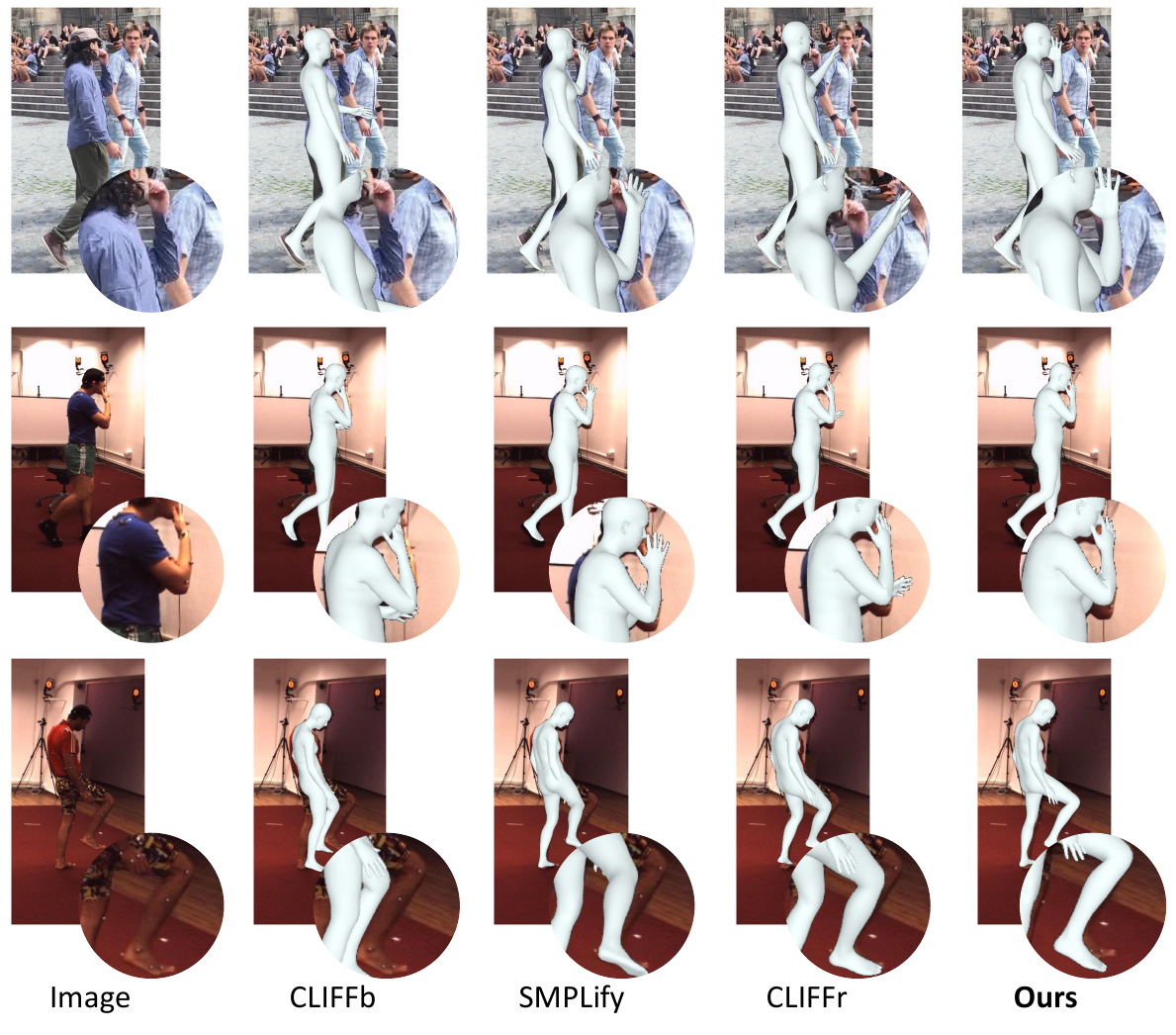}
    \vspace{-0.6cm}
	\caption{More comparison visualizations.}
	\label{fig:morecompare}
    \vspace{-0.25cm}
\end{figure*}

In this section, we present additional visualization results. Fig.~\ref{fig:moreloops} illustrates the refinement process over iterations on more examples, and Fig.~\ref{fig:morecompare} provides further comparisons with SMPLify~\cite{SMPLify_2016} and CLIFFr~\cite{CLIFF_2022}.

\section{Extra Discussion}
In this section, we provide more discussion about our method as follows

\textbf{1) Why previous methods are suboptimal in proximal joints?} \\
As discussed in the main paper, prior human refinement methods typically utilize parametric optimization to optimize all body joints collectively through gradient descent. However, this scheme has limitations with gradient descent: the gradient updates at different joints can be inconsistent or even conflicting. For instance, the optimal update directions for the Elbow and Wrist might significantly differ, as demonstrated in Fig.~\ref{fig:conflictgredient}. Such conflicts in gradient directions, when backpropagated to proximal joints like the Shoulder, can lead to complications in their updates, ultimately resulting in suboptimal outcomes in proximal joints refinement (as illustrated in Fig. \textcolor{red}{1b} of the main paper).

\textbf{2) What is the computation complexity of KITRO?} \\
As discussed in Sec. \textcolor{red}{4.3} of the main paper, the computation is efficient due to the decision tree's design, where the calculation depends solely on the depth, allowing nodes at each depth level to be processed in parallel. The most time-consuming part of our method is the Adam-based shape optimization and the total iteration number. However, Fig.~\ref{fig:shapeiter} and Fig. \textcolor{red}{5} (in the main paper) show that these numbers are both relatively low for decent results. In practice, KITRO completes testing on all 35,515 samples in 3DPW test set in 15 minutes on a single NVIDIA GeForce RTX 2080 Ti GPU. For comparison, under identical conditions, SMPLify requires 20 minutes, and CLIFFr requires extra fine-tuning on the whole model taking more than 10 hours.
% {
%     \small
%     \bibliographystyle{ieeenat_fullname}
%     \bibliography{main}
% }

\end{document}